\documentclass[journal]{IEEEtran}
\newtheorem{definition}{Definition}
\usepackage{cite}
\usepackage{amsmath,amssymb,amsfonts}
\usepackage{algorithmic}
\usepackage{graphicx}
\usepackage{textcomp}
\usepackage{xcolor}
\usepackage[numbers, sort&compress]{natbib}
\usepackage{mathrsfs}
\usepackage{verbatim}
\usepackage{booktabs}
\usepackage{amsmath}
\usepackage{url}
\usepackage{hyperref}

\newtheorem{property}{Property}

\newtheorem{proof}{Proof}[section]
\def\BibTeX{{\rm B\kern-.05em{\sc i\kern-.025em b}\kern-.08em
    T\kern-.1667em\lower.7ex\hbox{E}\kern-.125emX}}
\begin{document}

\title{IdentityDP: Differential Private Identification Protection for Face Images\\
\author{Yunqian~Wen,
        Li~Song,~\IEEEmembership{Senior Member,~IEEE,}
        Bo~Liu,~\IEEEmembership{Senior Member,~IEEE,}
        Ming~Ding,~\IEEEmembership{Senior Member,~IEEE,}
        and~Rong~Xie,~\IEEEmembership{Member,~IEEE}
\thanks{\noindent This work was submitted in part and published in the proceedings of IEEE International Conference on Visual Communications and Image Processing, 2020 \cite{wen2020hybrid}}
\IEEEcompsocitemizethanks{
\IEEEcompsocthanksitem Y. Wen, L. Song and R. Xie are with The Institute of Image Communication and Network Engineering, Shanghai Jiao Tong University, Shanghai 200240, China (email: wenyunqian@sjtu.edu.cn; song\_li@sjtu.edu.cn; xierong@sjtu.edu.cn).
\IEEEcompsocthanksitem B. Liu is with School of Computer Science, University of Technology Sydney, NSW 2007, Australia (email:bo.liu@uts.edu.au).
\IEEEcompsocthanksitem M. Ding is with Data61, Sydney, NSW, 1435 Australia (email:
ming.ding@data61.csiro.au)}       
}
}
\maketitle


\begin{abstract}
Because of the explosive growth of face photos as well as their widespread dissemination and easy accessibility in social media, the security and privacy of personal identity information becomes an unprecedented challenge. Meanwhile, the convenience brought by advanced identity-agnostic computer vision technologies is attractive. Therefore, it is important to use face images while taking careful consideration in protecting people's identities. Given a face image, face de-identification, also known as face anonymization, refers to generating another image with similar appearance and the same background, while the real identity is hidden. Although extensive efforts have been made, existing face de-identification techniques are either insufficient in photo-reality or incapable of well-balancing privacy and utility. In this paper, we focus on tackling these challenges to improve face de-identification. We propose IdentityDP, a face anonymization framework that combines a data-driven deep neural network with a differential privacy (DP) mechanism. This framework encompasses three stages: facial representations disentanglement, $\epsilon$-IdentityDP perturbation and image reconstruction. Our model can effectively obfuscate the identity-related information of faces, preserve significant visual similarity, and generate high-quality images that can be used for identity-agnostic computer vision tasks, such as detection, tracking, etc. Different from the previous methods, we can adjust the balance of privacy and utility through the privacy budget according to pratical demands and provide a diversity of results without pre-annotations. Extensive experiments demonstrate the effectiveness and generalization ability of our proposed anonymization framework.

\end{abstract}


\IEEEpeerreviewmaketitle

\section{Introduction}

\IEEEPARstart{T}{oday's} popularity of smartphones allows people to take their face photos conveniently. Particularly, the blooming development of media and network techniques makes a vast amount of photos more approachable. At the same time, however, advanced image retrieval and face verification models allow to index and examine privacy relevant information more reliably than ever. Consequently, among those image sources exposed to the public with or without our awareness, the wide range of private information inadvertently leaked is severely under-estimated \cite{PrivacyNet_TIP}.

Opportunities for misuse of the unprotected face image and advanced computer vision technologies are numerous and potentially disastrous \cite{Dual_TIP}. Restrictive laws and regulations such as the General Data Protection Regulations (GDPR) \cite{GDPR2018} has taken effect. GDPR requires regular consent from the individual for any use of their personal data to guarantee data privacy, however, it also makes the creation of high-quality datasets that include people becoming extremely challenging. Fortunately, if the data does not allow to identify the corresponding individual, entities are free to use the data without consent. what's more, many computer vision tasks in practice such as detection, tracking, or people counting, do not need to identify the people, but to detect them.

All the troubles and dilemmas mentioned above can be summarized to one issue: given a face image, how can we create another image with similar appearance and the same background, while the real identity is hidden and face detectors are still allowed to work? Traditional anonymization techniques are mainly obfuscation-based and always significantly alter the original face. Other previous work in this field is sparse and limited in both practicality and efficacy: \textit{k}-same algorithm-based methods \cite{newton2005preserving, gross2005integrating, gross2006model, du2014garp, jourabloo2015attribute} fail to make full use of existing data and deliver fairly poor visual quality; adversarial perturbation-based methods \cite{komkov2019advhat, oh2017adversarial, shafahi2018poison, liu2019protecting, zhu2019transferable, shan2020fawkes} usually depend highly on the accessibility of the target system and require special training; recent GAN-based methods\cite{li2019anonymousnet, wang2020infoscrub, sun2018natural, ren2018learning, hukkelaas2019deepprivacy, wu2019privacy, meden2017face, sun2018hybrid, blanz1999morphable, gafni2019live, maximov2020ciagan} have trouble generating visually similar de-identified faces as well. Note that there exists a trade-off between privacy protection and dataset utility \cite{Tradeoff_TIFS, hasan2018viewer}, and previous methods are unable to balance this matter.

\begin{figure*}[htbp]
\centerline{
\includegraphics[width=\textwidth]{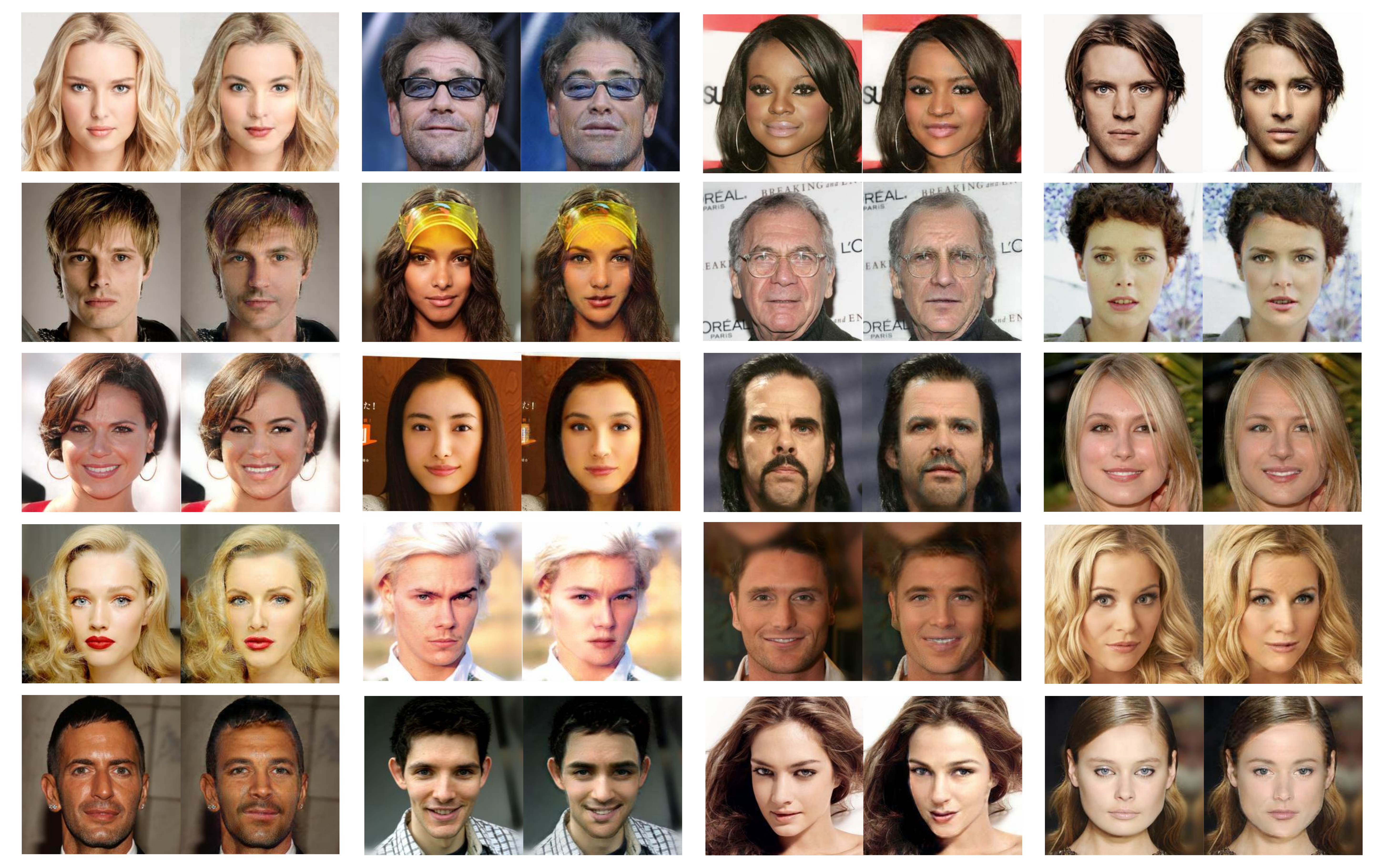}}
\caption{ IdentityDP for face anonymization. In each pair, left is the original image and right is the synthesized result with an altered identity. The results show that face identities are changed in a perceptually natural manner, and in the meantime, each pair of images still shares most of the non-identity related information. }
\label{show_n2}
\end{figure*}

To tackle these challenges, we propose IdentityDP, a framework that anonymize face images without significantly distorting the original images, nor destroying the availability of face detectors (see Fig. ~\ref{show_n2}). Especially, individuals are allowed to have control over the anonymization procedure to get the most suitable results in practice. IdentityDP achieves this by helping users adding well-designed obfuscation to photos' high-level identity representations. For example, a user who wants to share photos on social media or the public web can add adjustable perturbations according to his demands through our framework before uploading them. The uploaded photos will look similar to the original ones, but when an adversary employs a general face verificator to compare the user's face images with the altered ones, it will indicate that they are from different people.

The proposed IdentityDP framework consists of three stages. Stage-\uppercase\expandafter{\romannumeral1} aims to perform facial  representations disentanglement. We train a specially designed GAN for disentanglement between high-level identity representation and multi-level attribute representations in the feature space. Here the identity representation affects face verification systems to judge whether it is the same person, and the attribute representation guarantees the visual similarity. Stage-\uppercase\expandafter{\romannumeral2} carries out an $\epsilon$-IdentityDP mechanism, where adjustable differential privacy (DP) ~\cite{dwork2008differential} perturbations are applied to the identity representation. Stage-\uppercase\expandafter{\romannumeral3} implements the image reconstruction. In more detail, we fix the well-trained GAN network in Stage-\uppercase\expandafter{\romannumeral1}, and generate de-identified face images utilizing the perturbed identity representation as well as the original attribute representations. IdentityDP leverages both the GAN's outstanding ability to disentangle images' representations in the latent space and differential privacy theory, managing to balance the trade-off between image quality and privacy protection according to practical needs. In addition, our framework requires neither pre-annotation nor pre-detection of faces, but can generate numerous anonymous results.

Our contributions in this work are as follows:
\begin{itemize}
\item We propose a general framework that is suitable for the de-identification of people in face images.

\item As far as we know, we are the first to introduce the rigorously formulated DP theory into the face-anonymous task. The users are able to get not only high-quality anonymous images but also an adjustable privacy protection mechanism.

\item We demonstrate that our method does not require special training or targeted adjustments for many unauthorized identity verification systems or face datasets that never seen before.

\item We show that images anonymized by our method can be detected by common face detection models, so the processed images are still usable for identity-agnostic computer vision tasks (such as monitoring and tracking).

\item We show that our de-identified method is significantly less computationally complex and consumes a small amount of computing resources.
\end{itemize}

The remainder of this paper is organized as follows. In Section \uppercase\expandafter{\romannumeral2}, we summarize related work. Section \uppercase\expandafter{\romannumeral3} formalizes the face de-identification problem, introduces relevant DP theory and proposes our assumptions. Section \uppercase\expandafter{\romannumeral4} outlines the three-stages IdentityDP framework. Results of experiments analysing the proposed IdentityDP method and comparisons with existing methods are reported in Section \uppercase\expandafter{\romannumeral5}. we conclude in Section \uppercase\expandafter{\romannumeral6} with discussions of future research direction.

\section{Related Work}
In this section, we introduce the related work on face de-identification. We classify face anonymous methods into four categories: traditional obfuscation-based methods, \textit{k}-same algorithm-based methods, adversarial perturbation-based methods and GAN-based methods.

\subsection{Traditional Obfuscation-Based Methods}
In traditional computer vision studies, face de-identification technologies are mainly obfuscation-based. To be more specific, individuals can obfuscate privacy sensitive face area in an image by using approaches including blurring, pixelation, masking and so on. These traditional methods are widely used in daily life because of their simplicity and ease of operation. However, researchers have shown these techniques are vulnerable, and the private information in obfuscated images is still in danger of being leaked \cite{oh2016faceless}. McPherson \textit{et al.}\ \cite{mcpherson2016defeating} showed that deep learning methods especially CNN-based recognition models can successfully identify faces in images encrypted with these techniques with high accuracy. To make matters worse, obfuscation-based approaches towards manipulating images always tend to destroying the usability of images. Vishwamitra \textit{et al.}\ \cite{vishwamitra2017blur} indicated that both blurring and blocking would impact image perception scores, and even lower scores were observed for images obfuscated by blocking. Moreover, how to conduct a sufficient blur itself is non-trivial \cite{frome2009large}.

\subsection{\textit{k}-Same Algorithm-Based Methods}
To improve the performance of traditional methods, Newton \textit{et al.}\ \cite{newton2005preserving} introduced the first privacy-enabling algorithm, \textit{k}-same \cite{sweeney2002k}, to the context of image databases. By applying the \textit{k}-same algorithm, a given image is represented by an average face of \textit{k}-closet faces from the gallery. This procedure theoretically limits the performance of recognition to 1/\textit{k}, but the resulting images usually suffer from ghosting artifacts due to small alignment errors. Many variants of \textit{k}-same \cite{gross2005integrating, gross2006model, du2014garp, jourabloo2015attribute} were then proposed to improve the data utility and the naturalness of de-identified face images. Although these methods are once a mainstay of anonymous technology, they have notable limitations. Firstly, the \textit{k}-same assumes that each subject is only represented once in the datasets, but this may be violated in practice. The presence of multiple images from the same subject or images sharing similar biometric characteristics can lead to lower levels of privacy protection. Secondly, the k-same operates on a closed set and produces a corresponding de-identified set, which is not applicable in situations that involve processing individual images or sequences of images. Thirdly, their de-identified results always do not look natural enough, let alone resemble the original image. The above limitations indicate that there is still plenty of room for improvement in face de-identification research.

\subsection{Adversarial Perturbation-Based Methods}
New techniques and mechanisms are being applied to enhance image obfuscation. A fundamental idea is to generate a small but intentional worst-case disturbance to an original image, which misleads CNN-based recognition models without causing a significant difference perceptible to human eyes. Komkov and Petiushko \cite{komkov2019advhat} showed that carefully computed adversarial stickers on a hat could reduce its wearer's likelihood of being recognized. Oh \textit{et al.}\ \cite{oh2017adversarial} introduced a general framework based on game theory to conduct adversarial image perturbations and enforce guarantees on the user's level of privacy. An alternative to evading models is to disrupt their training via a data poisoning attack. Shafahi \textit{et al.}\ \cite{shafahi2018poison} presented an optimization-based method for crafting poison images, in which just one single poison image could control classifier behavior. Liu \textit{et al.} ~\cite{liu2019protecting} proposed to use adversarial perturbation to protect image privacy from both humans and AI. Zhu \textit{et al.}\ \cite{zhu2019transferable} introduced a new "polytope attack" in which poison images were designed to surround the targeted image in the feature space. Taking both ideas into account, Fawkes \cite{shan2020fawkes}, the state-of-the-art method, helped users wearing imperceptible "cloaks" to their own photos before releasing them. When used to train facial recognition models, these "cloaked" images produce functional models that consistently cause normal images of the user to be misidentified. Though their obfuscation performances are superb even at imperceptible perturbation level, these methods depend highly upon the accessibility to target systems, so can only be guaranteed for target-specific recognizers. In contrast, we hope to obfuscate identities against general face verification systems, and we are interested in gaining good generalization ability.

\subsection{GAN-Based Methods}
GANs represent an inspiring framework for generating sharp and realistic natural face image samples via a minimax game \cite{goodfellow2014generative}. It has therefore become popular in recent face de-identified techniques, which can be divided into three categories.

\noindent \textbf{Attribute manipulation-based methods.} Face attributes are crucial to face identification for human beings, and some methods achieve de-identification by manipulating attributes. Li \textit{et al.}\ \cite{li2019anonymousnet} proposed the Privacy-Preserving Attribute Selection (PPAS) algorithm to select and update facial attributes such that the distribution of any attribute was close to its real-life distribution, and provided measurable privacy for face anonymization with privacy guarantees. Wang \textit{et al.}\ \cite{wang2020infoscrub} introduced a bi-directional discriminator to alleviate issues of partial inversion of attributes, and executed attribute inversion and obfuscation in a two-stage manner.

\noindent \textbf{Conditional inpainting-based methods.} Since face is one of the strongest cues to infer a person's identity, 
a lot of studies cover up sensitive identity information by conditional inpainting face area. Sun \textit{et al.}\ \cite{sun2018natural} generated a realistic head inpainting based on 68 facial keypoints landmarks. Ren \textit{et al.}\ \cite{ren2018learning} trained a face modifier to remove privacy-sensitive information, while an action detector was trying to maximize spatial action detection performance. DeepPrivacy \cite{hukkelaas2019deepprivacy} directly removed the whole face area and generated new faces based on a sparse pose estimation, which ensured 100$\%$ removal of privacy-sensitive information in the original face. Wu \textit{et al.}\ \cite{wu2019privacy} designed a verificator to help remove biometric information and a regulator to maintain similar image utility. The involvement of these two types of prior knowledge was proved to significantly improve the model performance.

\noindent \textbf{Conditional ID-swapping-based methods.} Replacing the identity in a face image with someone else is a direct but effective idea of face anonymization. Meden \textit{et al.}\ \cite{meden2017face} proposed an de-identification pipeline that each generated face is a combination of $k$ identities. Sun \textit{et al.}\ \cite{sun2018hybrid} explicitly manipulated the identity through identity parameters provided by 3DMM \cite{blanz1999morphable}. Gafni \textit{et al.}\ \cite{gafni2019live} maximally decorrelated the identity conditioned on the high-level descriptor of a person's facial image, while having the perception (pose, illumination and expression) fixed. CIAGAN \cite{maximov2020ciagan} leveraged facial landmark and identity one hot-vector to remove the identification characteristics of people, while still keeping necessary features to allow face and body detectors to work.

Although GAN-based methods account for a substantial part of face de-identification study, they suffer from various conditional information requiring either manually annotations or computational resources, not to mention changed expressions, distorted shape, and loss of accessories. In this paper, we introduce a hybrid framework to try to solve the above problems.

\begin{figure*}[htbp]
\centerline{
\includegraphics[width=\textwidth]{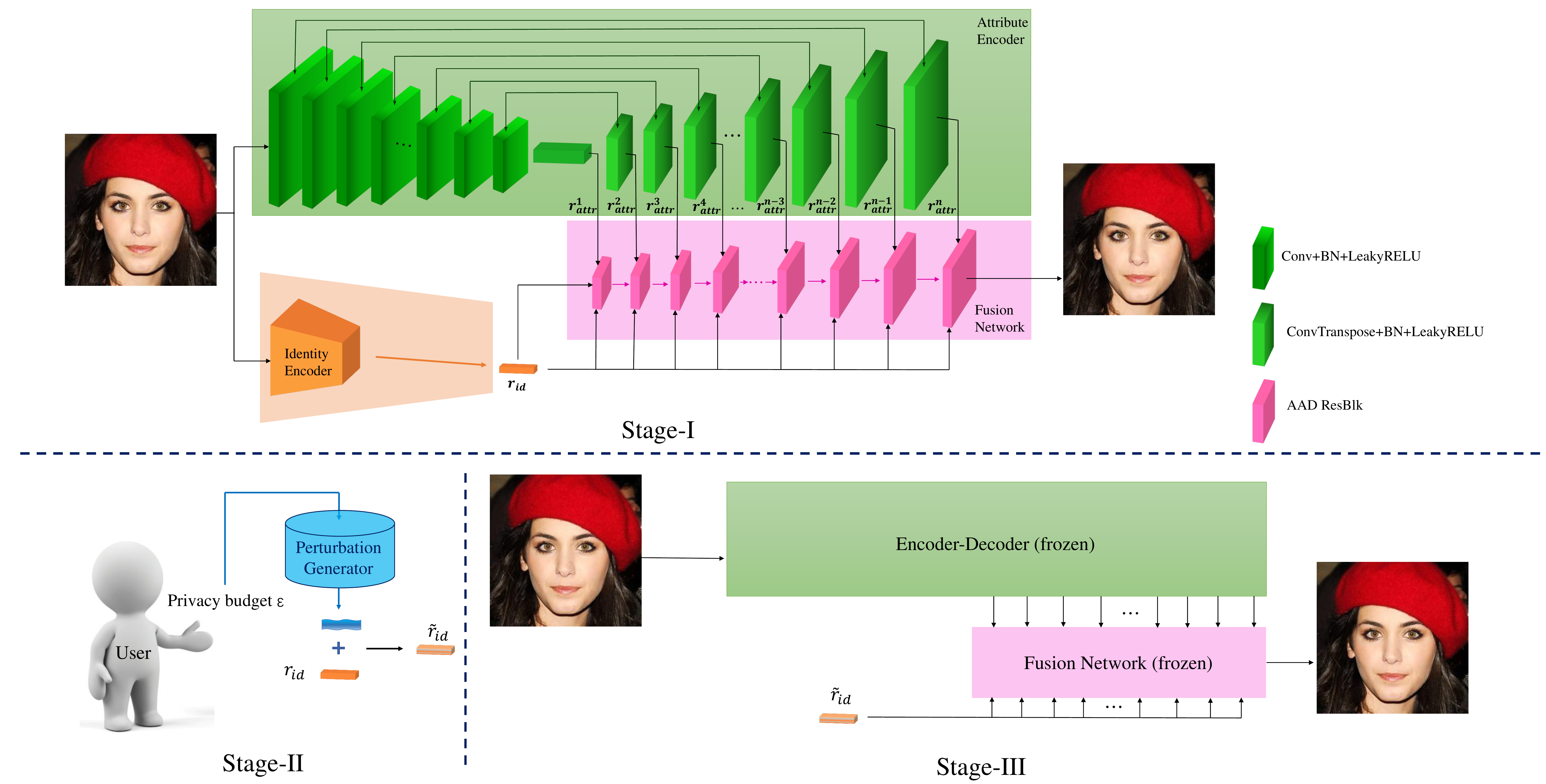}}
\caption{ Architecture of the proposed 3-stages IdentityDP framework, which based on a data-driven deep neural network and a Laplace $\epsilon$-IdentityDP mechanism. Stage-\uppercase\expandafter{\romannumeral1}: training a network to extract the disentangled high-level identity as well as attributes representations and restore the original faces; Stage-\uppercase\expandafter{\romannumeral2}: generating the perturbed identity representation under the Laplace $\epsilon$-IdentityDP mechanism; Stage-\uppercase\expandafter{\romannumeral3}: crafting anonymous faces from perturbed identity representation and original attribute representations through the frozen network. }
\label{network}
\end{figure*}

\section{Preliminaries}

\subsection{Problem Formulation}
A face de-identification model can be viewed as a transformation function $\delta$ that maps a given face image $X$ to a de-identified image $\hat{X}$, aiming to mislead face verification systems. Essentially, we are generating a new fake identity out of the input image. The problem can be formulated as follows:
\begin{gather}
\delta (X)= \hat{X} \\
s.t.: \text{Identity}\{ X \} \not= \text{Identity} \{ \hat{X} \}\nonumber.
\label{equ2}
\end{gather}
Meanwhile, considering image utility, $\hat{X}$ should look similar to $X$ as much as possible and be detectable by general face detectors.

\subsection{Differential Privacy Theory}

\subsubsection{Differential Privacy}
Differential Privacy (DP)~\cite{dwork2008differential}, a cryptography-inspired privacy-preserving model, guarantees that the likelihood of seeing an output on a given original datasets is close to the likelihood of seeing the same output on another datasets that differs from the original one in any single row. Here, the output could be another datasets, a statistical summary table, or a simple answer to a query, etc.
Generally speaking, the basic idea of a DP mechanism is to introduce randomness into the original datasets,
so that any individuals' information cannot be inferred by an adversary looking at the released output.

A formal definition of DP is shown below:
\begin{definition}{\label{def:DP}\textbf{($\epsilon$-Differential Privacy) }~\cite{Dwork2006}:}
A randomized mechanism $\mathcal{M}$ gives $\epsilon$-differential privacy if for any neighboring datasets $D$ and $D^\prime$ differing on one element, and all sets of output $S$:
\begin{align}
Pr[\mathcal{M}(D)\in S] \leq \exp(\epsilon) Pr[\mathcal{M}(D^\prime)\in S].
\end{align}
\end{definition}
This parameter $\epsilon$,
which is usually referred to as a privacy budget,
is a bound on the ratio of the likelihood probabilities of seeing the same output on neighbouring datasets.
The smaller the value of $\epsilon$,
the stronger the privacy guarantee.

A random perturbation can be added to achieve the differential privacy.
Sensitivity calibrates the amount of noise for a specified query $f$ of dataset $D$.
$\Delta f$ is the $l_1-$norm sensitivity defined as
\begin{definition}{\label{def: sensitivity}\textbf{($l_1-$norm sensitivity) }\cite{Dwork2006}:}
For any query $f$: $D \to \mathbb{R}$, $l_1-$norm sensitivity is the maximum $l_1-$ norm of $f(D)-f(D^\prime)$, i.e.,
\begin{align}
\Delta f = \max\limits_{D, D^\prime}||f(D)-f(D^\prime)||_1.
\label{equ: sensitivity}
\end{align}
\end{definition}

The Laplace mechanism is one of the most generic mechanism to guarantee differential privacy \cite{dwork2014algorithmic}.
\begin{definition}{\label{def: Laplace mechanism}\textbf{(Laplace Mechanism) }\cite{Dwork2006}:}
Given a function $f$: $D \to \mathbb{R}$, the following mechanism $\mathcal{M}$ provides the $\epsilon$-Differential Privacy:
\begin{align}
\mathcal{M}(D) = f(D) + Lap(\frac{\Delta f}{\epsilon}).
\end{align}
\end{definition}

\subsubsection{Local Differential Privacy}
In traditional DP setting, there is a trusted curator who applies carefully calibrated random noise to the real values returned for a particular query. However, in many practical scenarios, the curator might not be trustworthy. The data needs to be randomised without the global knowledge. Local differential privacy (LDP) \cite{kairouz2014extremal} is applicable to this case. It is considered to be a strong and rigorous notion of privacy that provides plausible deniability and deemed to be a state-of-the-art approach for privacy-preserving data collection and distribution.

\begin{definition}{\label{def:LDP}\textbf{($\epsilon$-LDP):}} A randomized mechanism $\mathcal{M}$ satisfies $\epsilon$-LDP, if for any two inputs $X , X^\prime$ and the set of all possible outputs $\mathcal{Y}$, $\mathcal{M}$ satisfies:
\begin{align}
Pr[\mathcal{M}(X) \in \mathcal{Y}] \leq e^\epsilon \cdot Pr[\mathcal{M}(X^\prime) \in \mathcal{Y}].
\end{align}
\end{definition}

And the sensitivity in this case equals to
\begin{align}
\Delta f = \max\limits_{X, X^\prime}||f(X)-f(X^\prime)||_1.
\label{equ: LDP-sensitivity}
\end{align}

\subsubsection{Two Important Properties}
Our approach relies on two key properties of DP.
First is the widely used parallel composition property when designing mechanisms:

\begin{property}{\label{(parallel composition property)}\textbf{(Parallel composition) }~\cite{zhu2017differential}:}
Suppose we have a set of privacy mechanisms M = \{$M_1$,\ldots,$M_m$\}, if each $M_i$ provides $\epsilon_i$ privacy guarantee on a disjointed subset of the entire dataset, M will provide (max\{$\epsilon_1$,\ldots,$\epsilon_m$\})-differential privacy.
\end{property}

Second is the well-known post-processing property:
\begin{property}{\label{(post-processing property)}\textbf{(Post-processing property) }~\cite{bun2016concentrated}:}
Any computation applied to the output of an ($\epsilon$,$\delta$)-DP algorithm remains ($\epsilon$,$\delta$)-DP.
\end{property}
For example,
averaging, rounding or any change to the output will not impact the privacy of the data.
This means that an analyst can conduct any data post-processing on a released DP dataset and cannot reduce its privacy guarantee.

\subsection{Face Verification and Our Assumptions}
The key idea of face verification is to develop effective representations in feature space for reducing intra-personal variations while enlarging inter-personal differences \cite{SFace_TIP}. The most ideal state is directly learning a mapping from face images to a compact feature space where distances precisely correspond to a measure of identity similarity. There are currently two main types of solutions: one is metric learning-based, and contrastive loss \cite{sun2016sparsifying}, center loss \cite{wen2016discriminative}, triplet loss~\cite{schroff2015facenet} are proposed to enhance the discrimination power of features; the other is angular margin-based, and many efforts \cite{liu2017sphereface, wang2018additive, wang2018cosface, deng2019arcface} about angle margin penalty have greatly improved the verification accuracy. To some extent, anonymization can be considered as a task to protect someone's identity representations from being correctly classified.

Here we have an assumption that identity representations of one person in different feature spaces are interrelated.
Once a face image's high-level representation in one feature space is disturbed into the wrong identity category, its identity representations in other feature spaces would also be classified incorrectly.

\section{The Proposed IdentityDP Framework}
For a given original clean face image $X$, our proposed IdentityDP framework can be used to generate its anonymous face images $\hat{X}$ in a controllable manner. We factor the face de-identification task into three stages.
In the first stage, we use a person's image as input and disentangle the latent space information into two main representations, namely identity and attribute. Among them, identity representation is modeled by embedding features through an encoder, while attribute representations are modeled by multi-level embedding features through a decoder, then the original face image is restored in an adaptively manner. In the second stage, we impose $\epsilon$-IdentityDP perturbations on identity representation according to practical demands. In the third stage, we freeze all the parameters of the network, and reconstruct anonymous face image with the perturbed identity representation. The overall architecture of the IdentityDP framework is shown in Fig. \ref{network}.

\subsection{Stage-\uppercase\expandafter{\romannumeral1}: Facial representations disentanglement}
In stage \uppercase\expandafter{\romannumeral1}, given an input face image, our goal is to represent the image using two disentangled representations, $r_{id}$ and $r_{attr}$.
$r_{id}$ is expected to contain all the information relevant to the identity, and $r_{attr}$ contains the rest of information carried by the image.
We investigate how to generate satisfactory face images with a specific disentanglement intention (i.e. identity and attribute) in mind. The key idea is to explicitly guide the generation process by an appropriate representation of that intention. 
Therefore, our network consists of 3 components: (1) Identity Encoder; (2) Attribute Encoder; (3) Fusion Generator.

\noindent \textbf{Identity Encoder}: As mentioned before, studies on face verification and recognition make arduous efforts for finding suitable face features that can reduce intra-personal variations while enlarge inter-personal differences, which is exactly our requirement for identity representation. Therefore, we choose a pretrained state-of-the-art face recognition model \cite{deng2019arcface} as identity encoder. The identity representation ${r}_{id}(X)$ is defined to be the last feature vector generated before the final FC layer, and denoted as:
\begin{equation}
r_{id}(X) = f(X).
\label{equ: r_id}
\end{equation}
\noindent \textbf{Attribute Encoder}: Attribute representation, which determines pose, expression, illumination, background and so on, intuitively carries more spatial information than identity. Johnson \textit{et al.}\ \cite{johnson2016perceptual} illustrate that low-level features tend to preserve image content and overall spatial structure, and high-level features tend to preserve color, texture, and exact shape. In order to preserve different level details, we employ multi-level feature maps to represent the attributes. In specific, we feed the input image $X$ into a U-Net-like structure, and then use the feature maps generated from the U-Net decoder as the attributes representations. More formally, we denote

\begin{equation}
{r}_{att}(X) = g(X) = \left\{r_{att}^1(X), r_{att}^2(X), \cdots, r_{att}^n(X)\right\},
\label{equ: attributes}
\end{equation}
where $r_{att}^k(X)$ represents the \textit{k}-th level feature map from the U-Net decoder, \textit{n} is the number of feature levels.

This attributes encoder does not require any artificial annotations, it extracts the attributes using self-supervised training: we require that the generated de-identified face $\hat{X}$ and the original face $X$ have the same attributes embedding. The loss function will be introduced later in Eq. (\ref{equ1}).

\noindent \textbf{Fusion Network}: After obtaining the disentangled identity and attribute representations, we would like to learn a way to integrate them to reproduce the original face image, which will be used in our subsequent steps. Through a simple trial, we find that direct feature concatenation can easily lead to blurry results and is not expected to be used. Fortunately, Li \textit{et al.}\cite{li2019faceshifter} used \textit{Adaptive Attentional Denormalization}\ (AAD) ResBlk to achieve remarkable feature integration in multiple feature levels. They argue that the attention mechanism with denormalizations make the effective regions of features more adaptive to adjust; this is an appealing property for fusion network, since identity and attribute representations can participate in synthesizing different parts of the face. We integrate $n$ AAD ResBlks to the body of our fusion network. As illustrated in Fig.~\ref{network}, in stage-\uppercase\expandafter{\romannumeral1}, after extracting the identity representation $r_{id}$, and encoding multi-level attribute feature maps ${r}_{att}$, the fusion generator integrates them through cascaded AAD ResBlks to restore the original face image $X$:
\begin{equation}
X = h(r_{id}, {r}_{att}).
\label{equ: origin_reconstruction}
\end{equation}
The training of h(.) will be discussed in the following sections.

\subsection{Stage-\uppercase\expandafter{\romannumeral2}: $\epsilon$-IdentityDP perturbation}
Stage-\uppercase\expandafter{\romannumeral2} generates the perturbed identity representation under a novel Laplace $\epsilon$-IdentityDP mechanism, which is defined as follows:

\begin{definition}{\label{def:LDP}\textbf{($\epsilon$-IdentityDP Mechanism):}} A randomized mechanism $\mathcal{M}$ satisfies $\epsilon$-IdentityDP, i.e. if for any two inputs face images $X , X^\prime$ and the set of all possible outputs $\mathcal{Y}$, $\mathcal{M}$ satisfies:
$Pr[\mathcal{M}(X) \in \mathcal{Y}] \leq e^\epsilon \cdot Pr[\mathcal{M}(X^\prime) \in \mathcal{Y}].$
For a face image $X$, if:
\begin{align}
f(X) = r_{id}(X),
\label{equ: r_id}
\end{align}
and
\begin{align}
\mathcal{M}(X) = r_{id}(X) + Lap(\frac{\Delta f}{\epsilon}) = \tilde{r}_{id},
\label{equ:epsilon-IdentityDP}
\end{align}
\end{definition}
We say that $\mathcal{M}(X)$ satisfies $\epsilon$-IdentityDP.

And the sensitivity is calculated as follows:
\begin{align}
\Delta f = \max\limits_{X, X^\prime}||r_{id}(X)-r_{id}(X^\prime)||_1.
\label{equ: LDP-sensitivity}
\end{align}

To achieve $\epsilon$-IdentityDP mechanism, we employ a noise generator to generate suitable Laplace noise whose size equals to the high-level identity representation according to specific privacy budget $\epsilon$. Then we directly add the noise on the identity representation from Stage-\uppercase\expandafter{\romannumeral1}, intending to obfuscate people's identity.

\subsection{Stage-\uppercase\expandafter{\romannumeral3}: Image reconstruction}
Stage-\uppercase\expandafter{\romannumeral3} is conditioned on the obfuscated identity representation from Stage-\uppercase\expandafter{\romannumeral2} and the original multi-level attribute features from Stage-\uppercase\expandafter{\romannumeral1}. In order to achieve good de-identified results, we freeze all the parameters of the well-trained fusion network in Stage-\uppercase\expandafter{\romannumeral1}, and generate anonymous face image $\hat{X}$ through the fusion network using obfuscated identity representation and multi-level attribute representations:
\begin{equation}
\hat{X} = h(\mathcal{M(X)}, g(X)) = h(\tilde{r}_{\text{id}}, {r}_{att}).
\label{equ: reconstruction}
\end{equation}

It can be approved that the generated image $\hat{X}$ follows $\epsilon$-IdeneityDP.
\begin{proof}
First, according to definition in Eq. (\ref{equ:epsilon-IdentityDP}), $\mathcal{M(X)}$ satisfies $\epsilon$-IdentityDP:
\begin{align}
&\frac{\textit{Pr}(\tilde{r}_{\text{id}}|f(X))}{\textit{Pr}(\tilde{r}_{\text{id}}|f(X^\prime))} = \prod_{i=1}^m \frac{\exp(-|r_{id(i)}-f(X)_i|/\frac{\Delta f}{\epsilon})}{\exp(-|r_{id(i)}-f(X^\prime)_i|/\frac{\Delta f}{\epsilon})} \notag \\
& = \prod_{i=1}^m \exp ( \frac{\epsilon(|r_{id(i)}-f(X^\prime)_i|-|r_{id(i)}-f(X)_i|)}{\Delta f}) \notag \\ &
\leq \prod_{i=1}^m \exp ( \frac{\epsilon|f(X)_i-f(X^\prime)_i|}{\Delta f}) \notag \\ &
=\exp(\frac{\epsilon \cdot \sum_{i=1}^m |f(X)_i-f(X^\prime)_i|}{\Delta f})\notag \\ &
=\exp(\frac{\epsilon \cdot \left \| f(X)-f(X^\prime) \right \|_1}{\Delta f})\notag \\&
\leq \exp(\epsilon), \notag
\end{align}
where the first inequality follows from that $\left|a\right|- \left|b\right| \leq \left|a-b\right|$ for any $a,b\in\mathbb{R}$. The rest of proof follows from the post-processing property of DP. Hence, we can conclude that if the identity representation is treated with DP noises, then the reconstructed face image $\hat{X}$ also satisfies the $\epsilon$-IdentityDP defined in Definition \ref{def:LDP}.
\end{proof}

\subsection{Training Process}
In Stage-\uppercase\expandafter{\romannumeral1}, we need to build a network which can not only disentangle identity and attribute representations, but also restore the original input face image from these two representations.

We utilize adversarial training for this framework. Let ${L}_{adv}$ be the adversarial loss for making $\hat{X}$ realistic. It is implemented as a multi-scale discriminator \cite{park2019semantic} on the downsampled output images:

\begin{equation}
{L}_{adv}(\hat{X},X) = log{D}_{img}(X) + log(1-{D}_{img}(\hat{X})).
\end{equation}

An identity preservation loss is used to preserve the identity of the source. It is formulated as:

\begin{equation}
{L}_{id} = 1 - \cos(r_{id}(\hat{X}), r_{id}(X)),
\end{equation}
where $\cos(\cdot , \cdot)$ represents the cosine similarity of two vectors. We also use the attributes preservation loss, which is formulated as:

\begin{equation}
{L}_{att} = \frac{1}{2}\sum_{k=1}^n \left\| r_{att}^k(\hat{X}) - r_{att}^k(X) \right\|_2^2.
\label{equ1}
\end{equation}

The reconstruction loss as pixel level L-2 distances between the target image $\hat{X}$ and X:

\begin{equation}
{L}_{rec} = \frac{1}{2} \left\| \hat{X} - X \right\|_2^2.
\end{equation}

The full objective to train our network in the first stage is a weighted sum of above losses as:

\begin{equation} \label{total}
{L}_{total} = {L}_{adv} + \lambda_{att} {L}_{att} + \lambda_{id} {L}_{id} + \lambda_{rec} {L}_{rec},
\end{equation}
where $\lambda_{att}$, $\lambda_{id}$ and $\lambda_{rec}$ are the weight parameters for balancing different terms.

In practice, GAN is hard to train, so adjusting the training strategy according to real-time generation effect is necessary. In order to use visualization tools to judge our training effect and make appropriate adjustments in time, we extract identity and attribute representations from two faces randomly sampled from the training dataset and then fuse them together in stage-\uppercase\expandafter{\romannumeral1}. It is worth noting the reconstruction loss should be set to ${L}_{rec} = 0$ when the two faces are different.

\section{Experiments}
\subsection{Experimental Setup}
1) \textit{Datasets}: We choose the CelebA-HQ datasets, which contains 30K high-resolution celebrity images with diverse demographic information like age, gender, and race~\cite{karras2017progressive}, to train our network in stage-\uppercase\expandafter{\romannumeral1}. We randomly select 27K images for training and 3K for testing. Moreover, in order to demonstrate our generalization ability and compare with conditional comparisons conveniently, we also test IdentityDP on the CelebA \cite{liu2015deep} datasets. All images are aligned and cropped to size $256 \times 256$ covering the whole face, as well as some background regions.

2) \textit{Comparison methods}: To validate the effectiveness of the proposed IdentityDP framework, we compare to traditional anonymization methods as well as state-of-the-art methods.

\begin{itemize}
\item Traditional Anonymization methods. We use Pixelization, Noise and Blur of faces.
\item State-of-the-art methods. We select 4 methods: AnonymousNet \cite{li2019anonymousnet}, DeepPrivacy \cite{hukkelaas2019deepprivacy}, CIAGAN \cite{maximov2020ciagan} and Fawkes \cite{shan2020fawkes}.
\end{itemize}

\subsection{Evaluation Metrics}
We evaluate all methods in privacy metrics as well as utility metrics.

1) Privacy metrics. Two different metrics are used to measure the performance of privacy protection.

\begin{itemize}
\item \textit{Identity Distance $\mathcal{ID\_DIS}$}. We employ FaceNet identification model \cite{schroff2015facenet} based on Inception-Resnet backbone, pre-trained on two public datasets: CASIA-Webface \cite{yi2014learning} and VGGFace2 \cite{cao2018vggface2}, whose LFW accuracy can reach 99.05\% and 99.65\% individually. The output distance of FaceNet can indicate the pairs of input faces' identity difference.
\item \textit{Protection success rate $\mathcal{PSR}$}. Besides publicly available datasets and known model architectures for academic usage, we also wish to understand the performance of IdentityDP on public facial verification services that people may touch in daily life. Therefore Microsoft Azure Face \cite{API} is employed to evaluate real-world effectiveness of a method. It gives judgement of whether the input pairs are of the same people. The protection success rate is the proportion of faces that are judged as different from the original ones.
\end{itemize}

2) Utility metrics. Two different metrics are used to evaluate the utility of processed images.

\begin{itemize}
\item \textit{PSNR} and \textit{SSIM}. We choose peak-signal-to-noise ratio (PSNR) as well as structural similarity index measure (SSIM) as two objective measures of similarity between anonymous results and original faces.
\item \textit{Face detection rate $\mathcal{FDR}$}. We evaluate whether the processed images are still usable for identity-agnostic computer vision tasks by performing face detection using HOG \cite{dalal2005histograms} Detector, and we calculate the proportion of faces that can be detected in the protected images.
\end{itemize}

\subsection{Implementation Details}
We implement our framework as shown in Fig.~\ref{network}. 
The number of attribute representation is set to $n=8$ (Eq. (\ref{equ: attributes})). In the training process, we use the Adam optimizer \cite{kingma2014adam} with momentum parameters $\beta_1 = 0, \beta_2 = 0.999$. The learning rate is set to 0.0004. The parameters in Eq. (\ref{total}) are set to $\lambda_{att} = \lambda_{rec} = 10, \lambda_{id} = 5$. 


\subsection{$\epsilon$-IdentityDP Mechanism Analysis}
To explicitly understand the differential privacy mechanism in our proposed IdentityDP, we design an experiment to explore how the privacy budget $\epsilon$ affects the face anonymization performance.
First of all, we extract every test image's identity representation and calculate the $l_1-$norm sensitivity $\Delta f$, i.e., $\Delta f = \max\limits_{X, X^\prime}||r_{id}(X)-r_{id}(X^\prime)||_1$, $X, X^\prime\in$ \textit{test datasets}.
Then we increase $\epsilon$ from 1.1 to 800, and accordingly adjust the IdentityDP framework.
Since our $\epsilon$-IdentityDP mechanism $\mathcal{M}(X)$ is $\mathcal{M}(X) = r_{id}(X) + Lap(\frac{\Delta f}{\epsilon})$, we double $\epsilon$ for better display effect and 100 anonymous faces are generated for every test face under each $\epsilon$.
Finally, various statistical mean metric values are calculated at each $\epsilon$ value.

\begin{figure}[htbp]
\centerline{
\includegraphics[width=\linewidth]{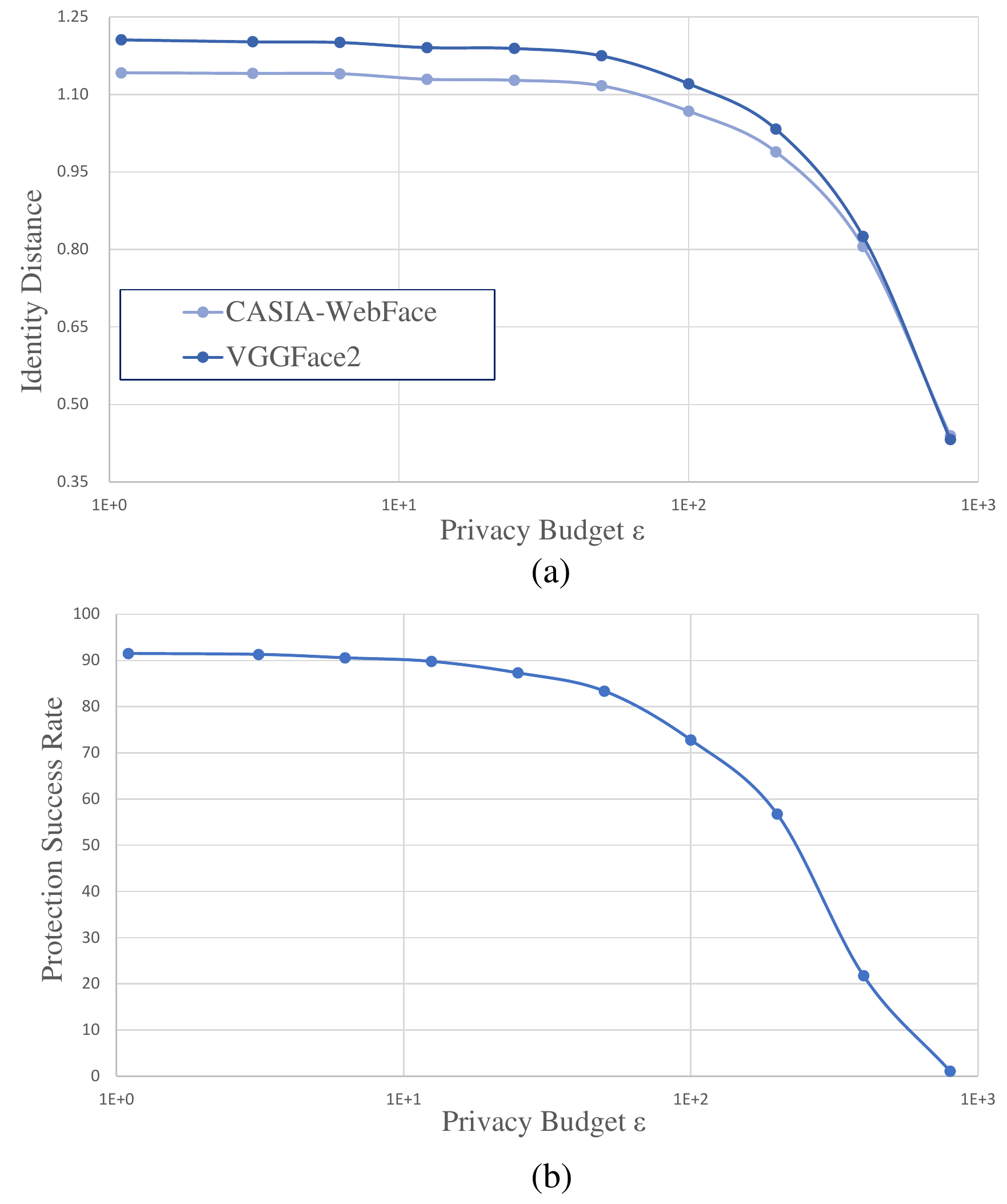}}
\caption{Identity protection performance: (a) the identity distance calculated by FaceNet model trained on CASIAWebface and VGGFace2 datasets respectively; (b) the Protection success rate calculated through public facial verification service \cite{API}. }
\label{privacy}
\end{figure}

\begin{figure}[htbp]
\centerline{
\includegraphics[width=\linewidth]{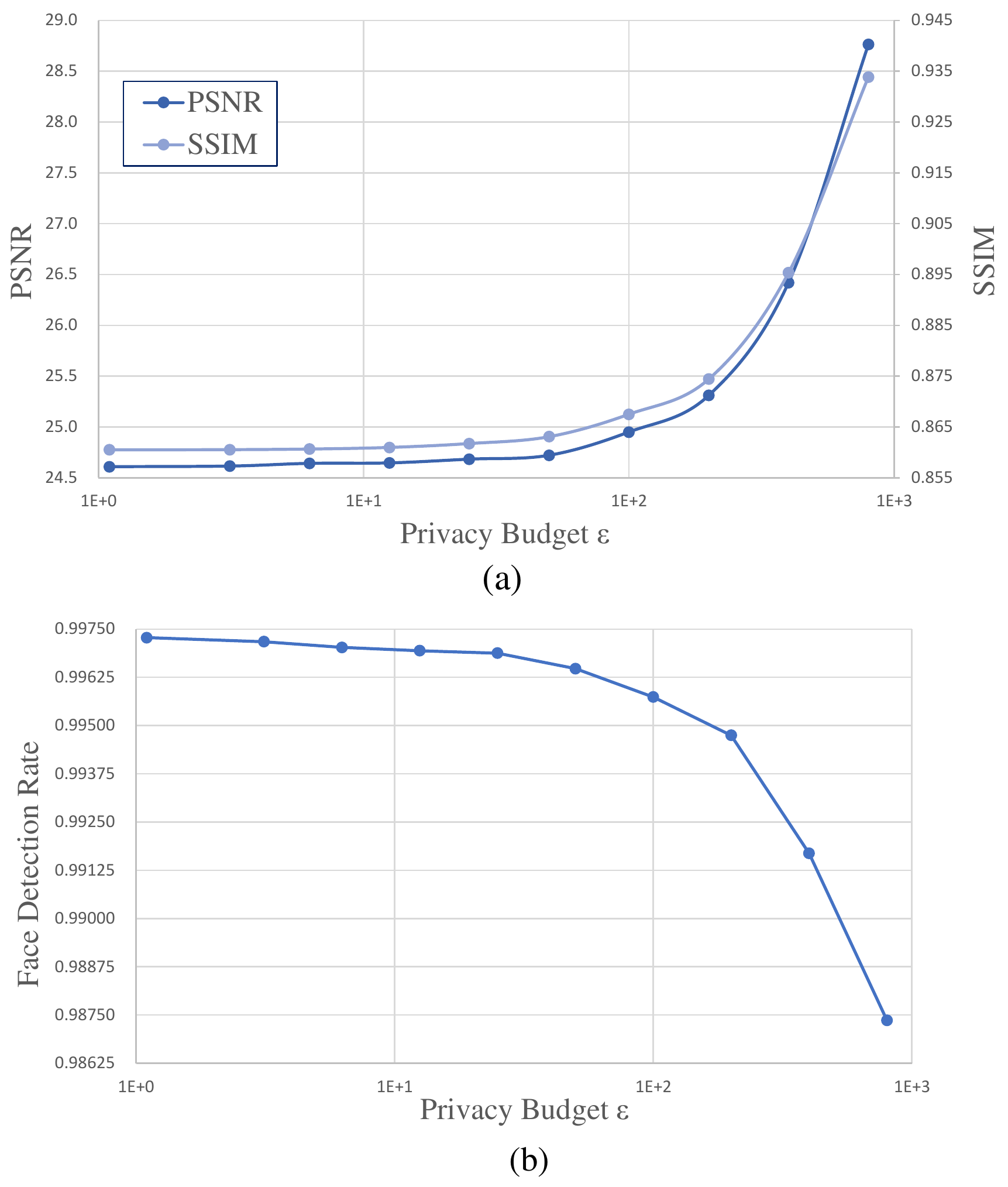}}
\caption{ Image utility performance: (a) PSNR and SSIM; (b) the Face detection rate calculated through HOG detector. }
\label{utility}
\end{figure}

\begin{figure*}[htbp]
\centerline{
\includegraphics[width=\textwidth]{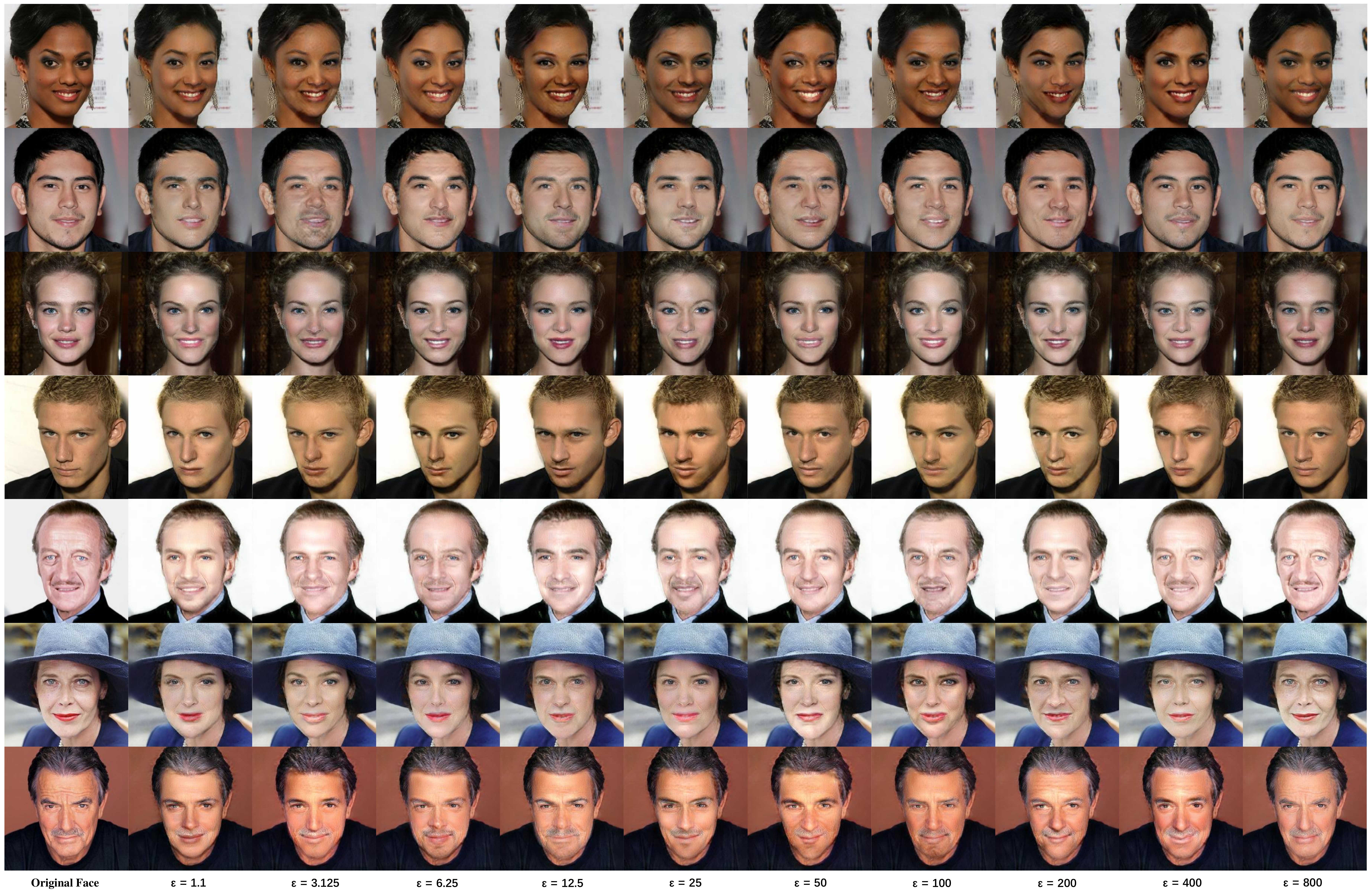}}
\caption{ Qualitative comparison of the influence of parameter $\epsilon$. The first column shows the original face images. The rest columns demonstrate anonymous face whose identity distance is closest to the mean distance under every $\epsilon$.}
\label{DP_self}
\end{figure*}

For privacy protection, when $\epsilon$ increase from 1.1 to 800, Fig. ~\ref{privacy} (a) shows that the average identity distance decreases gradually and Fig. ~\ref{privacy} (b) shows that the protection success rate decrease from 91.510\% to 1.125\%, illustrating that a smaller privacy budget guarantees better de-identified results. We show anonymous image whose identity distance is closest to the mean distance under every $\epsilon$ in Fig. ~\ref{DP_self} for visual observation, which also implies the diversity of our de-identified results.
For data utility, Fig. ~\ref{utility} (a) plots PSNR and SSIM vs. $\epsilon$, indicating that the visual similarity gets better as the privacy budget increases. Fig. ~\ref{utility} (b) shows that our face detection rate always remains at a high level, demonstrating that identity-agnostic computer vision technologies can still work on our processed faces.

Furthermore, an unexpected issue is that the face detection rate decreases slightly as $\epsilon$ increases. After research, we find the reason is that partially severely blocked faces in test datasets can recover some facial features in the blocked area using our framework, resulting in the detection of originally undetectable faces.

Based on a number of experiments, we get some experience in choosing a suitable privacy budget value: if the image's hue is light or the people's expression is exaggerated, a smaller privacy budget should be chosen. We recommend the user to set their privacy budget between 5 and 15 to obtain anonymous face efficiently. Specifically, our privacy budget is set to 6 during the subsequent experiments.

Fig. ~\ref{show_n2} illustrates some de-identified results in pairs, where left is the original image and right is the result generated by our framework. It demonstrates that human identities are obfuscated in a perceptually natural manner, in the meantime, each pair of images still shares similar appearance, as well as the same expression and background. It is worth noticing that our results can well retain the unique attributes of characters, such as rare hairstyles, beards, glasses and other accessories, which is hard to achieve in previous GAN-based methods.

\begin{table}[htbp]
\setlength{\abovedisplayskip}{1pt} 
\caption{QUANTITATIVE EVALUATION ON CELEBA-HQ DATASETs UNDER DIFFERENT METRICS}
\begin{center}
\resizebox{\linewidth}{15mm}{
\begin{tabular}{ccccccc}
\toprule
{}&{\centering $\mathcal{ID\_DIS}$(CASIA)}&{\centering $\mathcal{ID\_DIS}$(VGGFace2)}&{\centering $\mathcal{PSR}$}&{\centering PSNR}&{\centering SSIM}&{\centering $\mathcal{FDR}$} \\
\midrule
Pixelization($8\times8$) & 0.8646 & 0.8993 & 0 & 26.735 & 0.7671 & 0.923\\
Pixelization($16\times16$) & \textbf{1.1541} & \textbf{1.2195} & 0.017 & 23.926 & 0.7223 & 0.058\\
Noise($\sigma$ = 9) & 0.3317 & 0.2723 & 0.002 & 23.831 & 0.8312 & 0.986\\
Noise($\sigma$ = 49) & 1.1267 & 1.0280 & 0.012 & 14.370 & 0.5533 & 0.425\\
Blur($7\times7$) & 0.8491 & 0.8380 & 0 & 27.405 & 0.806 & 0.888\\
Blur($19\times19$) & 1.1102 & 1.1857 & 0.669 & 24.829 & 0.7719 & 0.518\\
DeepPrivacy & 1.0860 & 1.1829 & 0.961 & 21.012 & 0.7808 & 0.989\\
Fawkes & 0.7267 & 0.8585 & 0 & \textbf{35.898} & \textbf{0.9487} & 0.985\\
Ours($\epsilon$=6) & 1.1403 & 1.2012 & \textbf{0.908} & 24.640 & 0.8606 & \textbf{0.997}\\
\bottomrule
\end{tabular}}
\label{tab1}
\end{center}
\end{table}

\subsection{Comparisons with Traditional Methods}

In this subsection, the following traditional methods are implemented: (1) Pixelization: we cluster face region's pixels that are close in 2D space and color space, and then replace each cluster ($8\times8$, $16\times16$) with its average value. (2) Noise: we add Gaussian noise ($\sigma$ = 9, 49) on each pixel's RGB value of the face region; (3) Blur: following \textit{Ryoo et al.}\ \cite{ryoo2016privacy}, we downsample the face region to extreme low-resolution ($7\times7$, $19\times19$) and then upsample back. We set the privacy budget to 6. It can be seen that for the fairness of comparison, we select two parameters for each traditional method: one aims to make the identity distance close to our approach, at this time, the utility metrics are mainly compared; the other aims to make PSNR or SSIM close to our method, at this time, the privacy metrics are mainly compared.

\begin{figure*}[htbp]
\centerline{
\includegraphics[width=\textwidth]{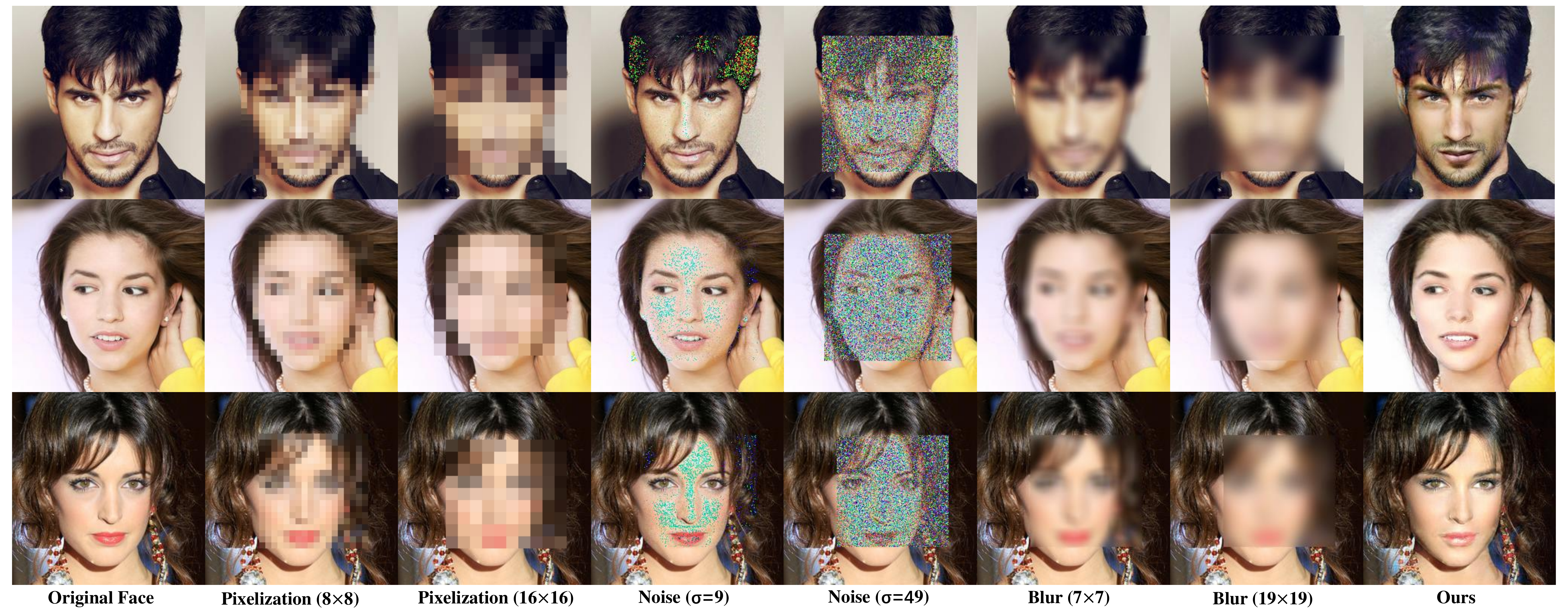}}
\caption{ Qualitative comparison with traditional methods. From left to right: original faces, faces obfuscated by Pixelization($4\times4$, $8\times8$), Noise($\sigma$ = 9, 18), Blur($8\times8$, $16\times16$), faces generated by our method.}
\label{traditonal}
\end{figure*}

Fig.~\ref{traditonal} shows the qualitative results. It is obvious that our approach achieves a great advantage in visual similarity as well as realism. The detailed quantitative results are shown in Table~\ref{tab1}, illustrating that the traditional methods fail to improve the privacy-utility trade-off and perform poorly in preventing practical face verification.

\subsection{Comparisons with State-of-the-art Methods}
In this subsection, we compare our IdentityDP with state-of-the-art face de-identification methods. Among them, DeepPrivacy and Fawkes are trained and tested on CelebA-HQ datasets. Anonymousnet and CIAGAN require pre-annotations and are trained on CelebA datasets, so we transfer our framework on CelebA and compare with them for fairness. We evaluate performance with these methods respectively.

1) \textit{Comparisons with Attribute manipulation-based Anonymization}: Facial attributes, including gender, age, haircut and so on, should be an important reference for identifying faces' identities, especially affecting human's subjective judgment. Therefore, manipulating face attributes to make faces anonymous seems reasonable. AnonymousNet, a privacy-preserving attribute selection algorithm for facial image obfuscation, is a typical representative.
\begin{figure}[htbp]
\centerline{
\includegraphics[width=\linewidth]{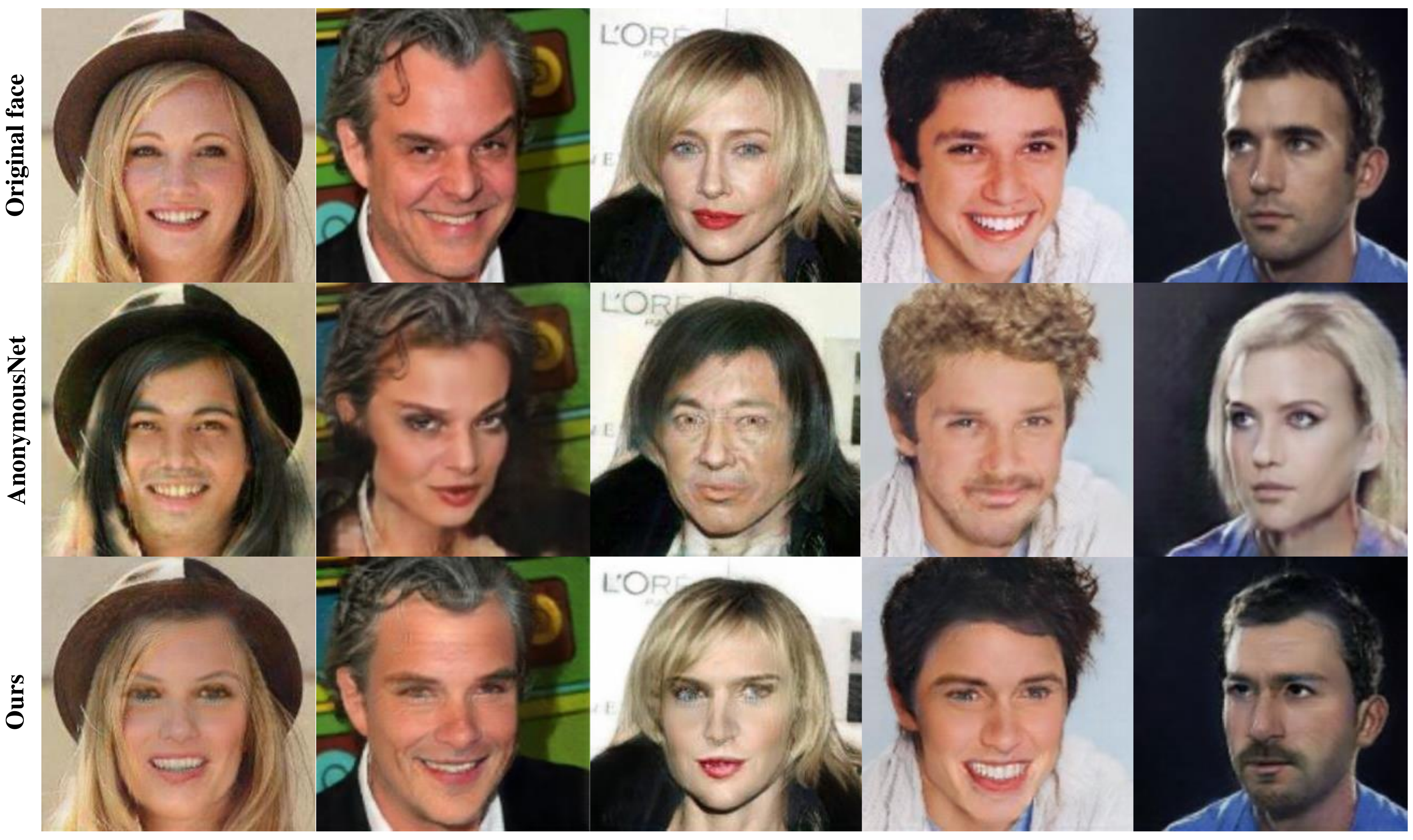}}
\caption{ Qualitative comparison of our method with AnonymousNet \cite{li2019anonymousnet}. The top row shows original faces, the second row shows corresponding anonymous faces generated by AnonymousNet, and the third row shows our results. }
\label{anonymousnet}
\end{figure}
Fig. ~\ref{anonymousnet} shows the anonymous faces generated from our framework and those from AnonymousNet. Due to the change of several face attributes, the anonymous face generated by AnonymousNet is often visually different from the original face, especially when modifying gender, which is not conducive to the normal use of the images. In contrast, our method achieves significant improvement in visual similarity. As can be seen from Table~\ref{tab2}, our method performs better under both privacy metrics and utility metrics, not to mention that AnonymousNet requires detailed data annotations. Moreover, it is worth noticing that although anonymous faces generated by AnonymousNet are visually very different from the original one, face verification service API can still judge them correctly, which suggests that general face attributes are not directly related to human identity.

\begin{table}[htbp]
\caption{QUANTITATIVE EVALUATION ON CELEBA DATASETS UNDER DIFFERENT METRICS}
\begin{center}
\resizebox{\linewidth}{7mm}{
\begin{tabular}{ccccccc}
\toprule
{}&{\centering $\mathcal{ID\_DIS}$(CASIA)}&{\centering $\mathcal{ID\_DIS}$(VGGFace2)}&{\centering $\mathcal{PSR}$}&{\centering PSNR}&{\centering SSIM}&{\centering $\mathcal{FDR}$} \\
\midrule
AnonymousNet & 0.8896 & 1.0589 & 0.295 & 18.892 & 0.7192 & 0.892\\
CIAGAN & 0.8155 & 1.0271 & \textbf{0.945} & 21.863 & 0.7401 & 0.958\\
Ours($\epsilon$=6) & \textbf{0.9345} & \textbf{1.0918} & 0.905 & \textbf{23.353} & \textbf{0.8188} & \textbf{0.986}\\
\bottomrule
\end{tabular}}
\label{tab2}
\end{center}
\end{table}

2) \textit{Comparisons with Conditional inpainting-based Anonymization}: Exposure of faces is the source of private information leakage. Therefore, some methods directly feed their networks with face-removing images as well as auxiliary annotations to automatically generate anonymous human faces. In this way, the generator never touches original faces, ensuring the removal of any privacy-sensitive information. DeepPrivacy is such a method which requires two annotations: a bounding box to identify the privacy-sensitive area and a sparse seven keypoints pose estimation of the face. It generates de-identified faces considering the original pose and image background. We compare our method with it.
\begin{figure}[htbp]
\centerline{
\includegraphics[width=\linewidth]{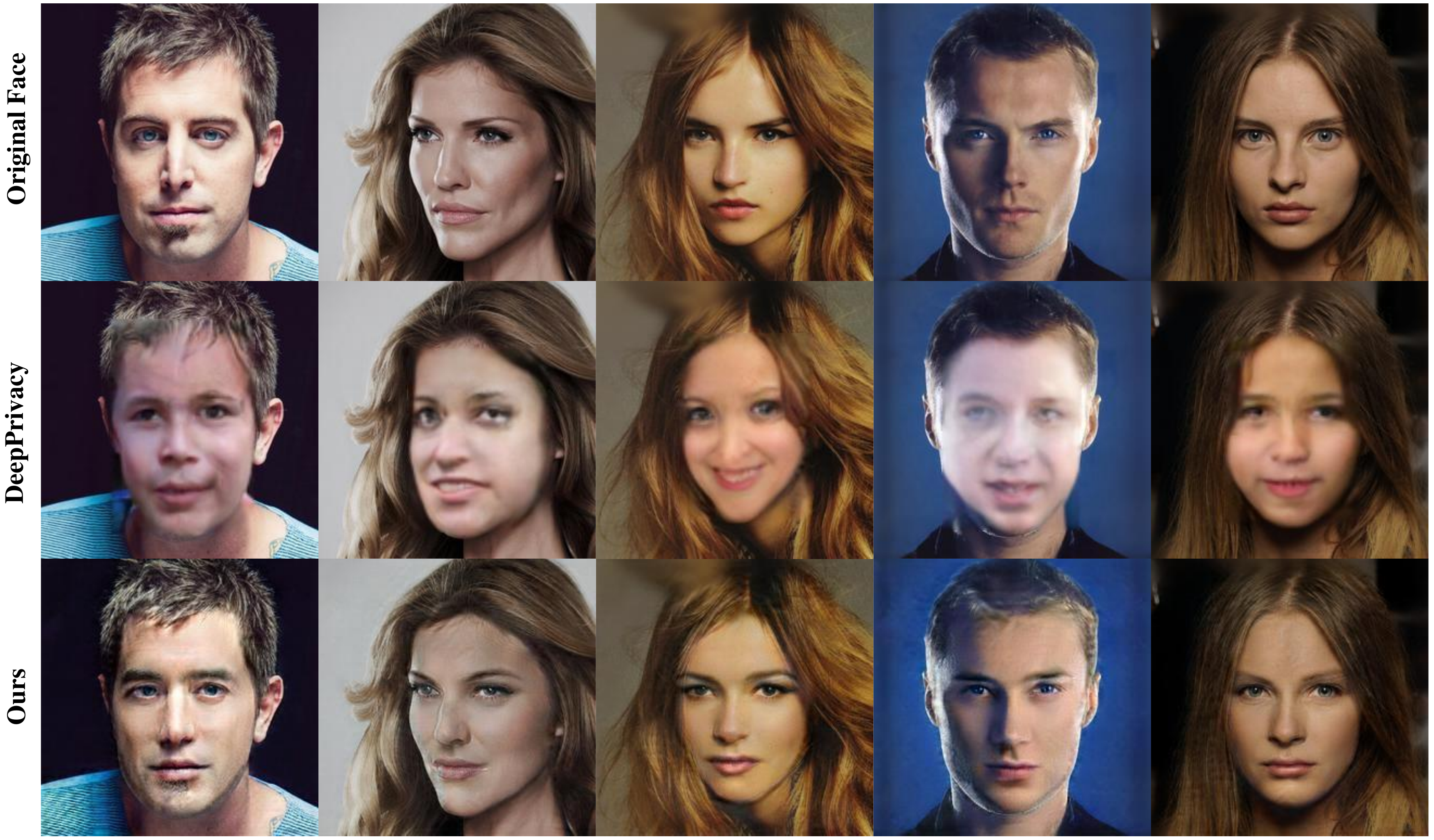}}
\caption{ Qualitative comparison of our method with DeepPrivacy \cite{hukkelaas2019deepprivacy}. The top row shows original faces, the second row shows corresponding anonymous faces generated by DeepPrivacy, and the third row shows our results. }
\label{DeepPrivacy}
\end{figure}
Fig. ~\ref{DeepPrivacy} reports the difference of methods. We can see that the face generated by DeepPrivacy can maintain the facial pose well, but is not visually similar to the original image. Besides, distortions and artifacts often occur. Our method produces more visual-pleasing anonymous faces which look similar to the original one. Table~\ref{tab1} shows quantitative results, our method is slightly inferior to DeepPrivacy in terms of privacy protection, but has remarkable data utility improvement.

3) \textit{Comparisons with Conditional ID-swapping-based Anonymization}: Since anonymizing a face is intended to hide its original identity, swapping the original ID with others may be a straightforward idea. Conditioned on face landmark and masked background image of the input image, CIAGAN generates a new fake identity out of the input image to achieve anonymization.

\begin{figure}[htbp]
\centerline{
\includegraphics[width=\linewidth]{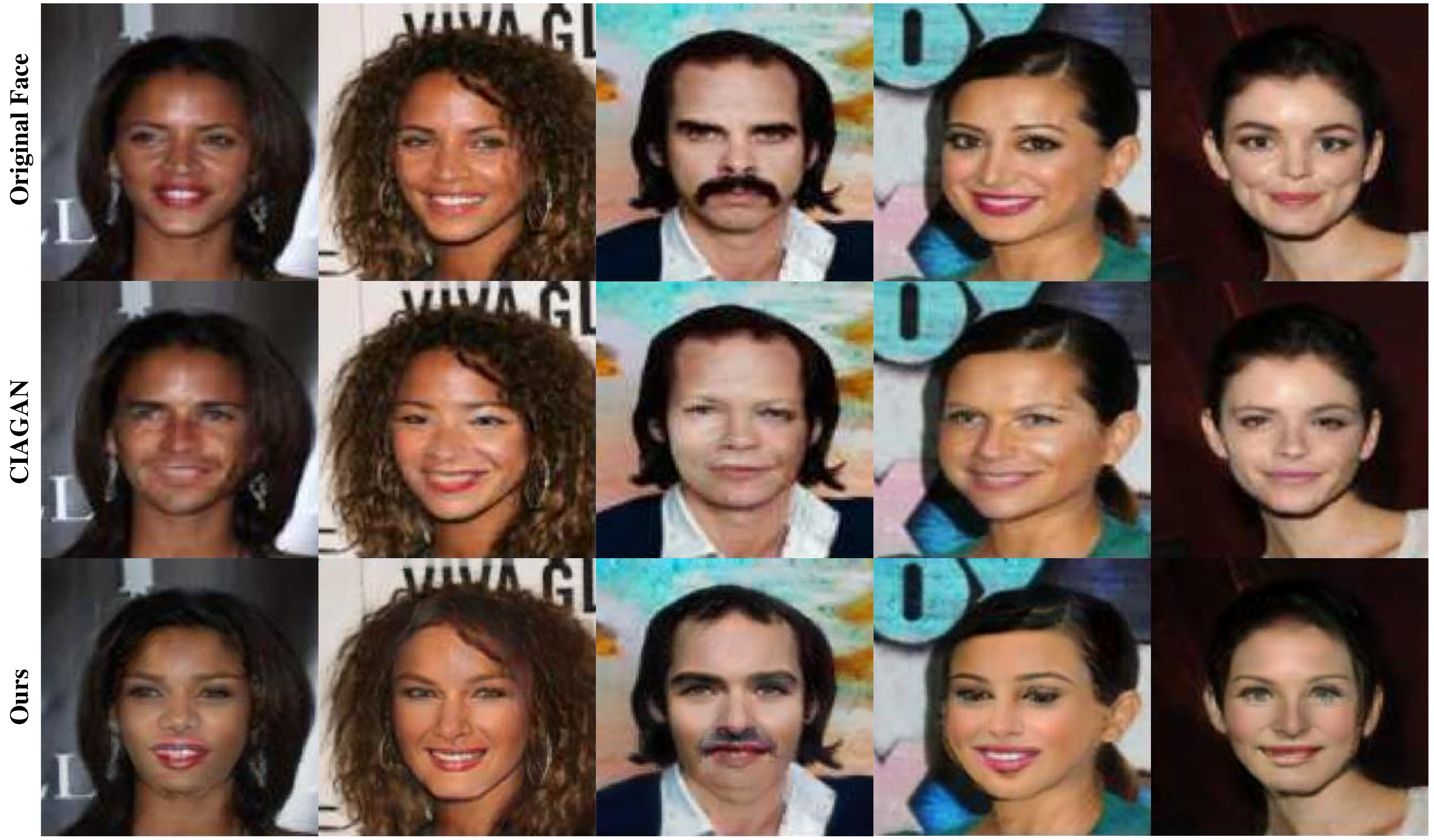}}
\caption{Qualitative comparison of our methods with CIAGAN \cite{maximov2020ciagan}. The top row shows original faces, the second row shows corresponding anonymous faces generated by CIAGAN, and the last row shows our results. }
\label{CIAGAN}
\end{figure}

We compare images generated from our proposed framework and those from \cite{maximov2020ciagan}. From Fig. ~\ref{CIAGAN} we can see that the two methods produce comparable results, while ours enjoy a better visual similarity. Table~\ref{tab2} shows quantitative results. In general, CIAGAN protects privacy better, and we maintain image utility better. However, CIAGAN has some notable flaws: 1) It needs to borrow someone else's identity as operation guidance, which may affect the privacy and security of the provider; 2) The effect of CIAGAN is brilliant only when the fake ID provider shares the same gender, similar age as well as similar makeup with the original people, which makes it inconvenient to use; 3) CIAGAN fails to maintain some special attributes, such as glasses, heavy makeup, and big beard; 4) CIAGAN depends on landmark detection to provide pre-annotations, which is troublesome and results in any face that has not been detected can not be anonymized. In contrast, our approach does not have these problems.

4) \textit{Comparisons with  Adversarial Perturbation-Based Anonymization}: De-identified methods based on adversarial examples are continuously popular because of their almost the same anonymous results. However, their performance depends largely on the accessibility of the target system's internal parameters, or special training on the target system. Fawkes, as one of the latest representatives, is selected as our comparison.

\begin{figure}[htbp]
\centerline{
\includegraphics[width=\linewidth]{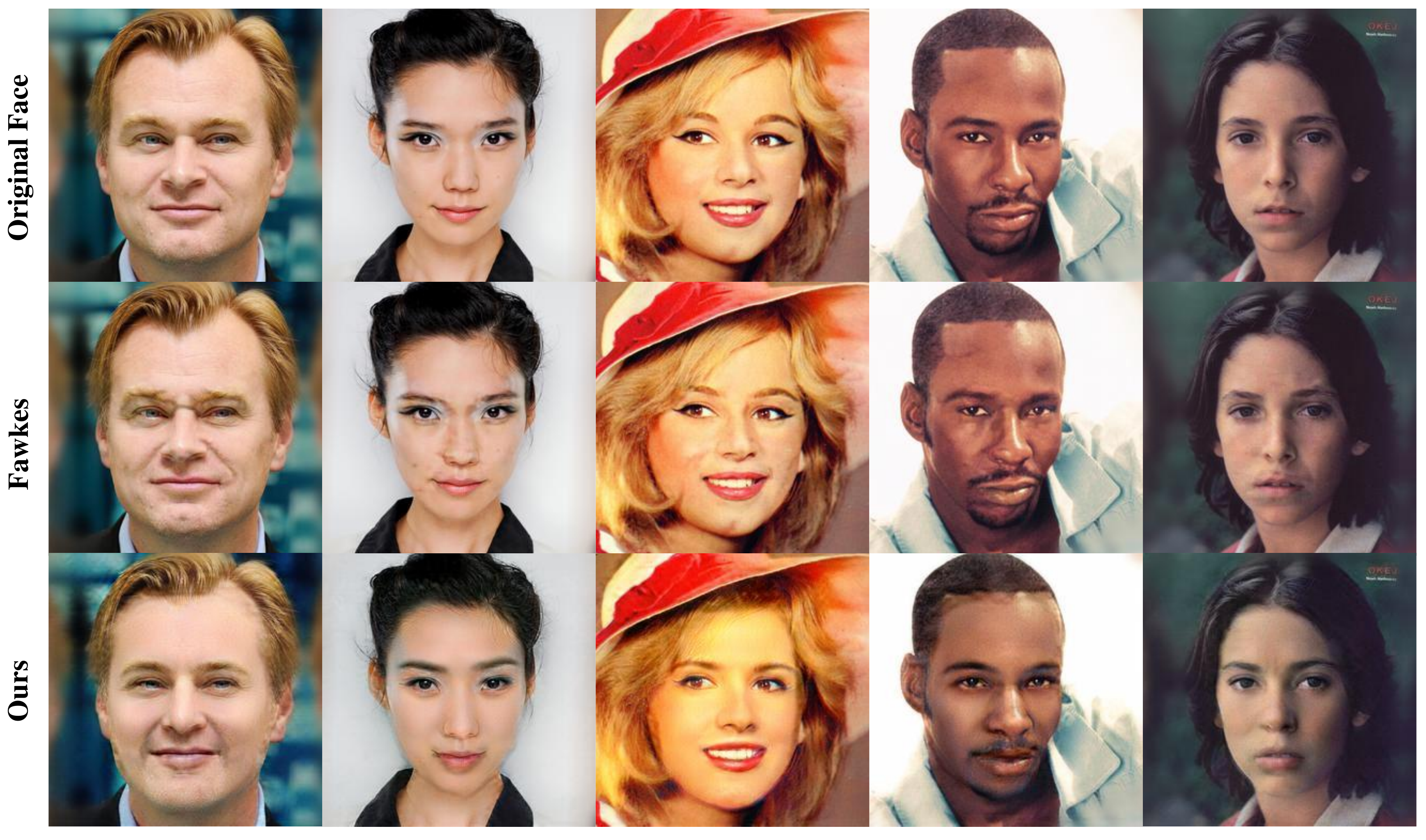}}
\caption{Qualitative comparison of our method with Fawkes \cite{shan2020fawkes}. The top row shows original faces, the second row shows corresponding anonymous faces generated by Fawkes, and the third row shows our results.}
\label{Fawkes}
\end{figure}

Fig. ~\ref{Fawkes} demonstrates that Fawkes can generate faces that look extremely like the original one, except for a few strange spots that sometimes appear. We just provide a comparable result. However, Table~\ref{tab1} shows that Fawkes performs poorly under privacy metrics, which means that faces processed by Fawkes are unable to obfuscate the previously inaccessible systems. In contrast, although our method suffers less visual similarity, it works better in preserving face privacy.

\subsection{Generalization Ability}
\begin{figure}[htbp]
\centerline{
\includegraphics[width=\linewidth]{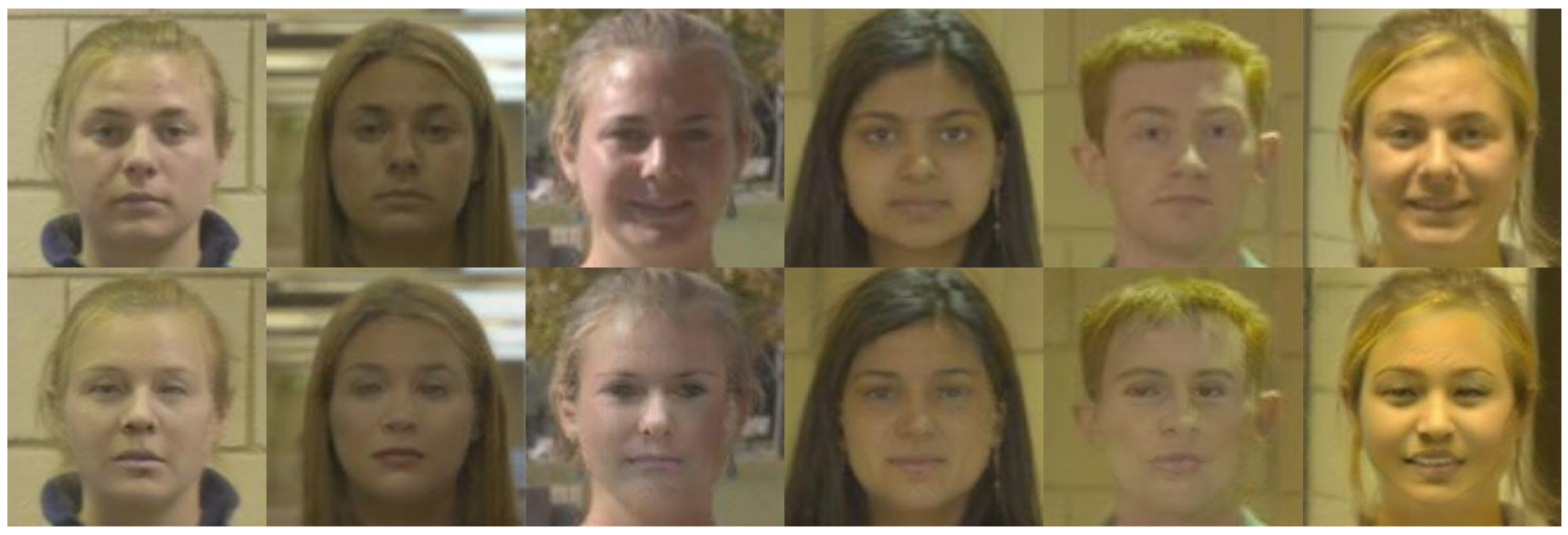}}
\caption{Our de-identification results on examples labeled as challenging or very challenging in the NIST Face Recognition Challenge \cite{phillips2011introduction}. The first row shows original faces, and the following row shows our corresponding de-identified results. }
\label{light}
\end{figure}

Our IdentityDP provides great generalization to various face images. In previous experiments, it has been proved by showing remarkable qualitative and quantitative results on CelebA, a datasets that our IdentityDP has never trained on before. To further demonstrate the robustness of our method, we apply our framework to face images from the very difficult inputs of \cite{phillips2011introduction}. As can be seen in Fig. ~\ref{light}, our method is robust to very challenging illuminations.

\begin{figure}[htbp]
\centerline{
\includegraphics[width=\linewidth]{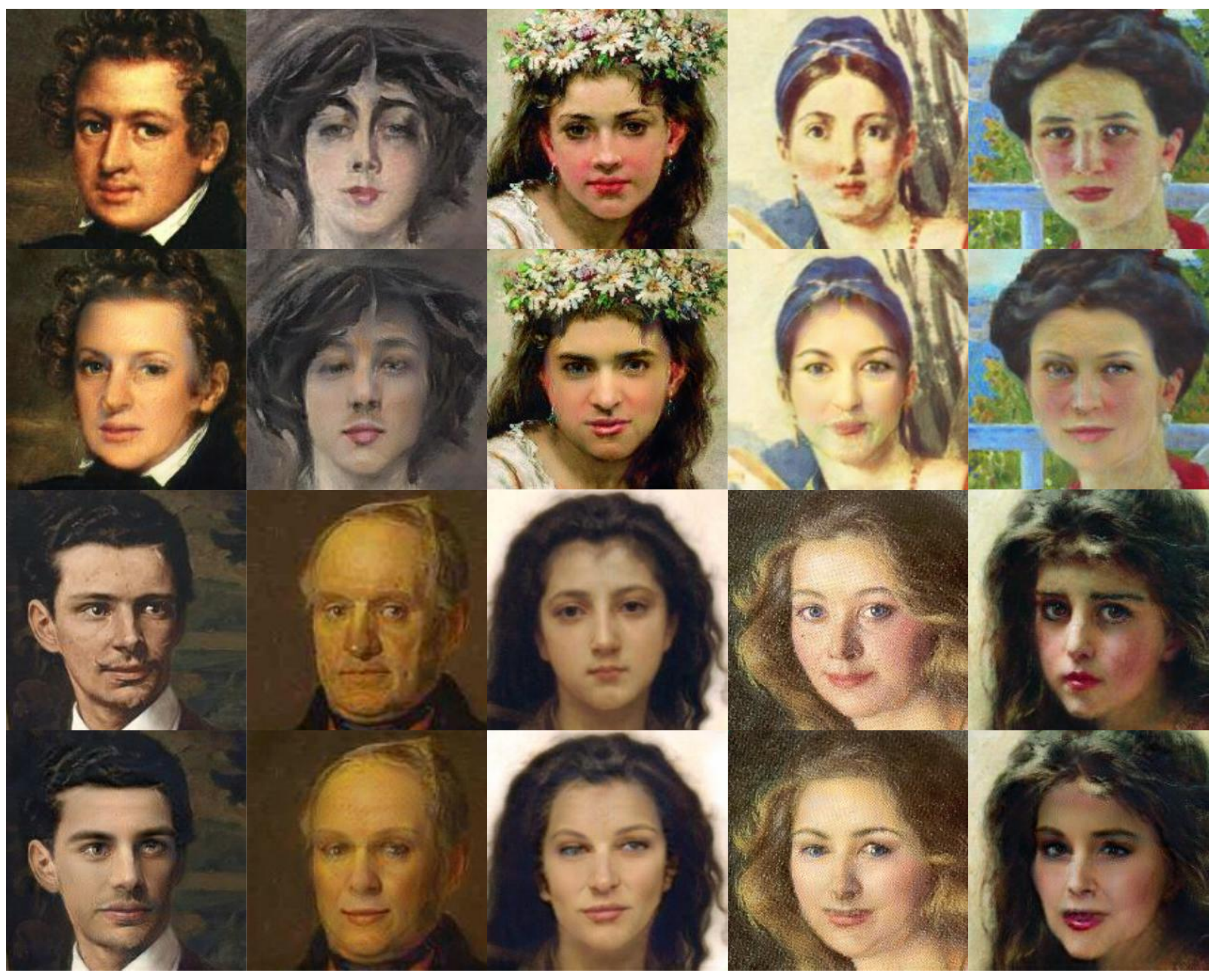}}
\caption{Our anonymization results on challenging artistic portraits. The first and the third row show the artistic portraits, while the second and the fourth row show our corresponding anonymous results. }
\label{portrait}
\end{figure}

In addition, we apply our framework on artistic portraits. All artworks are taken from Wikiart.org. Fig. ~\ref{portrait} shows the interesting results, illustrating that faces in different styles are anonymized successfully without causing significant distortions or artifacts.

\subsection{Computational Overhead}
In this subsection, we evaluate our computational overheads for anonymizing faces. IdentityDP adds little overhead for processing, as the only additions are a random noise tensor. On an NVIDIA GTX 1080 Ti, IdentityDP takes on average 0.329s per image. The low computational overhead is beneficial to process a large amount of face images.

\section{Conclusions and Future Work}
In this paper, we propose the IdentityDP framework that combines differential privacy mechanisms with deep neural networks to achieve image privacy protection for the first time. Our framework consists of three stages: deep representations disentanglement, $\epsilon$-IdentityDP perturbation and image reconstruction. In our framework, DP perturbation is directly added on to the identity representation to ensure privacy protection, while the attribute representation is unchanged and it preserves visual similarity well. Furthermore, the adjustable privacy budget guarantees the diversity of anonymization results. Experiments demonstrate the effectiveness of our framework in terms of privacy protection and image utility, and produce satisfactory results compared with the traditional as well as state-of-the-art methods. Moreover, our framework has a good generalization ability. In the future, we will further explore the trade-off between user privacy and authorized use of work. In addition, extending this work to videos and achieving temporal consistency would be an interesting direction.

\ifCLASSOPTIONcaptionsoff
  \newpage
\fi



\footnotesize
\bibliographystyle{IEEEtran}
\bibliography{mybibfilenew}

\begin{thebibliography}{10}
\providecommand{\url}[1]{#1}
\csname url@samestyle\endcsname
\providecommand{\newblock}{\relax}
\providecommand{\bibinfo}[2]{#2}
\providecommand{\BIBentrySTDinterwordspacing}{\spaceskip=0pt\relax}
\providecommand{\BIBentryALTinterwordstretchfactor}{4}
\providecommand{\BIBentryALTinterwordspacing}{\spaceskip=\fontdimen2\font plus
\BIBentryALTinterwordstretchfactor\fontdimen3\font minus
  \fontdimen4\font\relax}
\providecommand{\BIBforeignlanguage}[2]{{%
\expandafter\ifx\csname l@#1\endcsname\relax
\typeout{** WARNING: IEEEtran.bst: No hyphenation pattern has been}%
\typeout{** loaded for the language `#1'. Using the pattern for}%
\typeout{** the default language instead.}%
\else
\language=\csname l@#1\endcsname
\fi
#2}}
\providecommand{\BIBdecl}{\relax}
\BIBdecl

\bibitem{wen2020hybrid}
Y.~Wen, B.~Liu, R.~Xie, Y.~Zhu, J.~Cao, and L.~Song, ``A hybrid model for
  natural face de-identiation with adjustable privacy,'' in \emph{2020 IEEE
  International Conference on Visual Communications and Image Processing
  (VCIP)}.\hskip 1em plus 0.5em minus 0.4em\relax IEEE, 2020, pp. 269--272.

\bibitem{PrivacyNet_TIP}
V.~{Mirjalili}, S.~{Raschka}, and A.~{Ross}, ``Privacynet: Semi-adversarial
  networks for multi-attribute face privacy,'' \emph{IEEE Transactions on Image
  Processing}, vol.~29, pp. 9400--9412, 2020.

\bibitem{Dual_TIP}
I.~{Barron}, H.~J. {Yeh}, K.~{Dinesh}, and G.~{Sharma}, ``Dual modulated qr
  codes for proximal privacy and security,'' \emph{IEEE Transactions on Image
  Processing}, vol.~30, pp. 657--669, 2021.

\bibitem{GDPR2018}
E.~Commission, ``2018 reform of eu data protection rules,'' 2018.

\bibitem{newton2005preserving}
E.~M. Newton, L.~Sweeney, and B.~Malin, ``Preserving privacy by de-identifying
  face images,'' \emph{IEEE transactions on Knowledge and Data Engineering},
  vol.~17, no.~2, pp. 232--243, 2005.

\bibitem{gross2005integrating}
R.~Gross, E.~Airoldi, B.~Malin, and L.~Sweeney, ``Integrating utility into face
  de-identification,'' in \emph{International Workshop on Privacy Enhancing
  Technologies}.\hskip 1em plus 0.5em minus 0.4em\relax Springer, 2005, pp.
  227--242.

\bibitem{gross2006model}
R.~Gross, L.~Sweeney, F.~De~la Torre, and S.~Baker, ``Model-based face
  de-identification,'' in \emph{2006 Conference on computer vision and pattern
  recognition workshop (CVPRW'06)}.\hskip 1em plus 0.5em minus 0.4em\relax
  IEEE, 2006, pp. 161--161.

\bibitem{du2014garp}
L.~Du, M.~Yi, E.~Blasch, and H.~Ling, ``Garp-face: Balancing privacy protection
  and utility preservation in face de-identification,'' in \emph{IEEE
  International Joint Conference on Biometrics}.\hskip 1em plus 0.5em minus
  0.4em\relax IEEE, 2014, pp. 1--8.

\bibitem{jourabloo2015attribute}
A.~Jourabloo, X.~Yin, and X.~Liu, ``Attribute preserved face
  de-identification,'' in \emph{2015 International conference on biometrics
  (ICB)}.\hskip 1em plus 0.5em minus 0.4em\relax IEEE, 2015, pp. 278--285.

\bibitem{komkov2019advhat}
S.~Komkov and A.~Petiushko, ``Advhat: Real-world adversarial attack on arcface
  face id system,'' \emph{arXiv preprint arXiv:1908.08705}, 2019.

\bibitem{oh2017adversarial}
S.~J. Oh, M.~Fritz, and B.~Schiele, ``Adversarial image perturbation for
  privacy protection a game theory perspective,'' in \emph{2017 IEEE
  International Conference on Computer Vision (ICCV)}.\hskip 1em plus 0.5em
  minus 0.4em\relax IEEE, 2017, pp. 1491--1500.

\bibitem{shafahi2018poison}
A.~Shafahi, W.~R. Huang, M.~Najibi, O.~Suciu, C.~Studer, T.~Dumitras, and
  T.~Goldstein, ``Poison frogs! targeted clean-label poisoning attacks on
  neural networks,'' in \emph{Advances in Neural Information Processing
  Systems}, 2018, pp. 6103--6113.

\bibitem{liu2019protecting}
B.~Liu, J.~Xiong, Y.~Wu, M.~Ding, and C.~M. Wu, ``Protecting multimedia privacy
  from both humans and ai,'' in \emph{2019 IEEE International Symposium on
  Broadband Multimedia Systems and Broadcasting (BMSB)}.\hskip 1em plus 0.5em
  minus 0.4em\relax IEEE, 2019, pp. 1--6.

\bibitem{zhu2019transferable}
C.~Zhu, W.~R. Huang, A.~Shafahi, H.~Li, G.~Taylor, C.~Studer, and T.~Goldstein,
  ``Transferable clean-label poisoning attacks on deep neural nets,''
  \emph{arXiv preprint arXiv:1905.05897}, 2019.

\bibitem{shan2020fawkes}
S.~Shan, E.~Wenger, J.~Zhang, H.~Li, H.~Zheng, and B.~Y. Zhao, ``Fawkes:
  Protecting privacy against unauthorized deep learning models,'' in \emph{29th
  $\{$USENIX$\}$ Security Symposium ($\{$USENIX$\}$ Security 20)}, 2020, pp.
  1589--1604.

\bibitem{li2019anonymousnet}
T.~Li and L.~Lin, ``Anonymousnet: Natural face de-identification with
  measurable privacy,'' in \emph{Proceedings of the IEEE Conference on Computer
  Vision and Pattern Recognition Workshops}, 2019, pp. 0--0.

\bibitem{wang2020infoscrub}
H.-P. Wang, T.~Orekondy, and M.~Fritz, ``Infoscrub: Towards attribute privacy
  by targeted obfuscation,'' \emph{arXiv preprint arXiv:2005.10329}, 2020.

\bibitem{sun2018natural}
Q.~Sun, L.~Ma, S.~Joon~Oh, L.~Van~Gool, B.~Schiele, and M.~Fritz, ``Natural and
  effective obfuscation by head inpainting,'' in \emph{Proceedings of the IEEE
  Conference on Computer Vision and Pattern Recognition}, 2018, pp. 5050--5059.

\bibitem{ren2018learning}
Z.~Ren, Y.~Jae~Lee, and M.~S. Ryoo, ``Learning to anonymize faces for privacy
  preserving action detection,'' in \emph{Proceedings of the European
  Conference on Computer Vision (ECCV)}, 2018, pp. 620--636.

\bibitem{hukkelaas2019deepprivacy}
H.~Hukkel{\aa}s, R.~Mester, and F.~Lindseth, ``Deepprivacy: A generative
  adversarial network for face anonymization,'' in \emph{International
  Symposium on Visual Computing}.\hskip 1em plus 0.5em minus 0.4em\relax
  Springer, 2019, pp. 565--578.

\bibitem{wu2019privacy}
Y.~Wu, F.~Yang, Y.~Xu, and H.~Ling, ``Privacy-protective-gan for privacy
  preserving face de-identification,'' \emph{Journal of Computer Science and
  Technology}, vol.~34, no.~1, pp. 47--60, 2019.

\bibitem{meden2017face}
B.~Meden, R.~C. Mall{\i}, S.~Fabijan, H.~K. Ekenel, V.~{\v{S}}truc, and
  P.~Peer, ``Face deidentification with generative deep neural networks,''
  \emph{IET Signal Processing}, vol.~11, no.~9, pp. 1046--1054, 2017.

\bibitem{sun2018hybrid}
Q.~Sun, A.~Tewari, W.~Xu, M.~Fritz, C.~Theobalt, and B.~Schiele, ``A hybrid
  model for identity obfuscation by face replacement,'' in \emph{Proceedings of
  the European Conference on Computer Vision (ECCV)}, 2018, pp. 553--569.

\bibitem{blanz1999morphable}
V.~Blanz and T.~Vetter, ``A morphable model for the synthesis of 3d faces,'' in
  \emph{Proceedings of the 26th annual conference on Computer graphics and
  interactive techniques}, 1999, pp. 187--194.

\bibitem{gafni2019live}
O.~Gafni, L.~Wolf, and Y.~Taigman, ``Live face de-identification in video,'' in
  \emph{Proceedings of the IEEE International Conference on Computer Vision},
  2019, pp. 9378--9387.

\bibitem{maximov2020ciagan}
M.~Maximov, I.~Elezi, and L.~Leal-Taix{\'e}, ``Ciagan: Conditional identity
  anonymization generative adversarial networks,'' in \emph{Proceedings of the
  IEEE/CVF Conference on Computer Vision and Pattern Recognition}, 2020, pp.
  5447--5456.

\bibitem{Tradeoff_TIFS}
B.~{Rassouli} and D.~{Gündüz}, ``Optimal utility-privacy trade-off with total
  variation distance as a privacy measure,'' \emph{IEEE Transactions on
  Information Forensics and Security}, vol.~15, pp. 594--603, 2020.

\bibitem{hasan2018viewer}
R.~Hasan, E.~Hassan, Y.~Li, K.~Caine, D.~J. Crandall, R.~Hoyle, and A.~Kapadia,
  ``Viewer experience of obscuring scene elements in photos to enhance
  privacy,'' in \emph{Proceedings of the 2018 CHI Conference on Human Factors
  in Computing Systems}, 2018, pp. 1--13.

\bibitem{dwork2008differential}
C.~Dwork, ``Differential privacy: A survey of results,'' in \emph{International
  conference on theory and applications of models of computation}.\hskip 1em
  plus 0.5em minus 0.4em\relax Springer, 2008, pp. 1--19.

\bibitem{oh2016faceless}
S.~J. Oh, R.~Benenson, M.~Fritz, and B.~Schiele, ``Faceless person recognition:
  Privacy implications in social media,'' in \emph{European Conference on
  Computer Vision}.\hskip 1em plus 0.5em minus 0.4em\relax Springer, 2016, pp.
  19--35.

\bibitem{mcpherson2016defeating}
R.~McPherson, R.~Shokri, and V.~Shmatikov, ``Defeating image obfuscation with
  deep learning,'' \emph{arXiv preprint arXiv:1609.00408}, 2016.

\bibitem{vishwamitra2017blur}
N.~Vishwamitra, B.~Knijnenburg, H.~Hu, Y.~P. Kelly~Caine \emph{et~al.}, ``Blur
  vs. block: Investigating the effectiveness of privacy-enhancing obfuscation
  for images,'' in \emph{Proceedings of the IEEE Conference on Computer Vision
  and Pattern Recognition Workshops}, 2017, pp. 39--47.

\bibitem{frome2009large}
A.~Frome, G.~Cheung, A.~Abdulkader, M.~Zennaro, B.~Wu, A.~Bissacco, H.~Adam,
  H.~Neven, and L.~Vincent, ``Large-scale privacy protection in google street
  view,'' in \emph{2009 IEEE 12th international conference on computer
  vision}.\hskip 1em plus 0.5em minus 0.4em\relax IEEE, 2009, pp. 2373--2380.

\bibitem{sweeney2002k}
L.~Sweeney, ``k-anonymity: A model for protecting privacy,''
  \emph{International Journal of Uncertainty, Fuzziness and Knowledge-Based
  Systems}, vol.~10, no.~05, pp. 557--570, 2002.

\bibitem{goodfellow2014generative}
I.~Goodfellow, J.~Pouget-Abadie, M.~Mirza, B.~Xu, and D.~Warde-Farley,
  ``Generative adversarial nets in advances in neural information processing
  systems (nips),'' 2014.

\bibitem{Dwork2006}
C.~Dwork, F.~McSherry, K.~Nissim, and A.~Smith, ``Calibrating noise to
  sensitivity in private data analysis,'' in \emph{Theory of cryptography
  conference}.\hskip 1em plus 0.5em minus 0.4em\relax Springer, 2006, pp.
  265--284.

\bibitem{dwork2014algorithmic}
C.~Dwork, A.~Roth \emph{et~al.}, ``The algorithmic foundations of differential
  privacy.'' \emph{Foundations and Trends in Theoretical Computer Science},
  vol.~9, no. 3-4, pp. 211--407, 2014.

\bibitem{kairouz2014extremal}
P.~Kairouz, S.~Oh, and P.~Viswanath, ``Extremal mechanisms for local
  differential privacy,'' \emph{Advances in neural information processing
  systems}, vol.~27, pp. 2879--2887, 2014.

\bibitem{zhu2017differential}
T.~Zhu, G.~Li, W.~Zhou, and S.~Y. Philip, \emph{Differential privacy and
  applications}.\hskip 1em plus 0.5em minus 0.4em\relax Springer, 2017.

\bibitem{bun2016concentrated}
M.~Bun and T.~Steinke, ``Concentrated differential privacy: Simplifications,
  extensions, and lower bounds,'' in \emph{Theory of Cryptography
  Conference}.\hskip 1em plus 0.5em minus 0.4em\relax Springer, 2016, pp.
  635--658.

\bibitem{SFace_TIP}
Y.~{Zhong}, W.~{Deng}, J.~{Hu}, D.~{Zhao}, X.~{Li}, and D.~{Wen}, ``Sface:
  Sigmoid-constrained hypersphere loss for robust face recognition,''
  \emph{IEEE Transactions on Image Processing}, vol.~30, pp. 2587--2598, 2021.

\bibitem{sun2016sparsifying}
Y.~Sun, X.~Wang, and X.~Tang, ``Sparsifying neural network connections for face
  recognition,'' in \emph{Proceedings of the IEEE Conference on Computer Vision
  and Pattern Recognition}, 2016, pp. 4856--4864.

\bibitem{wen2016discriminative}
Y.~Wen, K.~Zhang, Z.~Li, and Y.~Qiao, ``A discriminative feature learning
  approach for deep face recognition,'' in \emph{European conference on
  computer vision}.\hskip 1em plus 0.5em minus 0.4em\relax Springer, 2016, pp.
  499--515.

\bibitem{schroff2015facenet}
F.~Schroff, D.~Kalenichenko, and J.~Philbin, ``Facenet: A unified embedding for
  face recognition and clustering,'' in \emph{Proceedings of the IEEE
  conference on computer vision and pattern recognition}, 2015, pp. 815--823.

\bibitem{liu2017sphereface}
W.~Liu, Y.~Wen, Z.~Yu, M.~Li, B.~Raj, and L.~Song, ``Sphereface: Deep
  hypersphere embedding for face recognition,'' in \emph{Proceedings of the
  IEEE conference on computer vision and pattern recognition}, 2017, pp.
  212--220.

\bibitem{wang2018additive}
F.~Wang, J.~Cheng, W.~Liu, and H.~Liu, ``Additive margin softmax for face
  verification,'' \emph{IEEE Signal Processing Letters}, vol.~25, no.~7, pp.
  926--930, 2018.

\bibitem{wang2018cosface}
H.~Wang, Y.~Wang, Z.~Zhou, X.~Ji, D.~Gong, J.~Zhou, Z.~Li, and W.~Liu,
  ``Cosface: Large margin cosine loss for deep face recognition,'' in
  \emph{Proceedings of the IEEE Conference on Computer Vision and Pattern
  Recognition}, 2018, pp. 5265--5274.

\bibitem{deng2019arcface}
J.~Deng, J.~Guo, N.~Xue, and S.~Zafeiriou, ``Arcface: Additive angular margin
  loss for deep face recognition,'' in \emph{Proceedings of the IEEE Conference
  on Computer Vision and Pattern Recognition}, 2019, pp. 4690--4699.

\bibitem{johnson2016perceptual}
J.~Johnson, A.~Alahi, and L.~Fei-Fei, ``Perceptual losses for real-time style
  transfer and super-resolution,'' in \emph{European conference on computer
  vision}.\hskip 1em plus 0.5em minus 0.4em\relax Springer, 2016, pp. 694--711.

\bibitem{li2019faceshifter}
L.~Li, J.~Bao, H.~Yang, D.~Chen, and F.~Wen, ``Faceshifter: Towards high
  fidelity and occlusion aware face swapping,'' \emph{arXiv preprint
  arXiv:1912.13457}, 2019.

\bibitem{park2019semantic}
T.~Park, M.-Y. Liu, T.-C. Wang, and J.-Y. Zhu, ``Semantic image synthesis with
  spatially-adaptive normalization,'' in \emph{Proceedings of the IEEE
  Conference on Computer Vision and Pattern Recognition}, 2019, pp. 2337--2346.

\bibitem{karras2017progressive}
T.~Karras, T.~Aila, S.~Laine, and J.~Lehtinen, ``Progressive growing of gans
  for improved quality, stability, and variation,'' \emph{arXiv preprint
  arXiv:1710.10196}, 2017.

\bibitem{liu2015deep}
Z.~Liu, P.~Luo, X.~Wang, and X.~Tang, ``Deep learning face attributes in the
  wild,'' in \emph{Proceedings of the IEEE international conference on computer
  vision}, 2015, pp. 3730--3738.

\bibitem{yi2014learning}
D.~Yi, Z.~Lei, S.~Liao, and S.~Z. Li, ``Learning face representation from
  scratch,'' \emph{arXiv preprint arXiv:1411.7923}, 2014.

\bibitem{cao2018vggface2}
Q.~Cao, L.~Shen, W.~Xie, O.~M. Parkhi, and A.~Zisserman, ``Vggface2: A dataset
  for recognising faces across pose and age,'' in \emph{2018 13th IEEE
  International Conference on Automatic Face \& Gesture Recognition (FG
  2018)}.\hskip 1em plus 0.5em minus 0.4em\relax IEEE, 2018, pp. 67--74.

\bibitem{API}
``Microsoft azure face api,'' \url{https://azure.microsoft.com/en-us/services/
  cognitive-services/face/.}

\bibitem{dalal2005histograms}
N.~Dalal and B.~Triggs, ``Histograms of oriented gradients for human
  detection,'' in \emph{2005 IEEE computer society conference on computer
  vision and pattern recognition (CVPR'05)}, vol.~1.\hskip 1em plus 0.5em minus
  0.4em\relax IEEE, 2005, pp. 886--893.

\bibitem{kingma2014adam}
D.~P. Kingma and J.~Ba, ``Adam: A method for stochastic optimization,''
  \emph{arXiv preprint arXiv:1412.6980}, 2014.

\bibitem{ryoo2016privacy}
M.~S. Ryoo, B.~Rothrock, C.~Fleming, and H.~J. Yang, ``Privacy-preserving human
  activity recognition from extreme low resolution,'' \emph{arXiv preprint
  arXiv:1604.03196}, 2016.

\bibitem{phillips2011introduction}
P.~J. Phillips, J.~R. Beveridge, B.~A. Draper, G.~Givens, A.~J. O'Toole, D.~S.
  Bolme, J.~Dunlop, Y.~M. Lui, H.~Sahibzada, and S.~Weimer, \emph{An
  introduction to the good, the bad, \& the ugly face recognition challenge
  problem}.\hskip 1em plus 0.5em minus 0.4em\relax IEEE, 2011.

\end{thebibliography}

%




\end{document}